%% file: Symmetries_of_Swarms_of_Mobile_Robots.tex
\documentclass[a4paper,UKenglish,cleveref, autoref, thm-restate]{lipics-v2021}

\pdfoutput=1 
\hideLIPIcs  

\usepackage{amsmath}

\usepackage{csquotes}

\title{Analyzing Symmetries of Swarms of Mobile Robots Using Equivariant Dynamical Systems} 

\titlerunning{Symmetries of Swarms of Mobile Robots} 

\author{Raphael Gerlach}{Paderborn University, Institute of Mathematics, 
Germany}{raphael.gerlach@uni-paderborn.de}{0009-0002-4750-2051}{}
\author{Sören von der Gracht}{Paderborn University, Institute of Mathematics, 
Germany}{soeren.von.der.gracht@uni-paderborn.de}{0000-0002-8054-2058}{Deutsche Forschungsgemeinschaft (DFG, German Research Foundation)--–453112019.}

\authorrunning{R. Gerlach and S. v. d. Gracht}

\Copyright{Raphael Gerlach and Sören von der Gracht}

\ccsdesc{Theory of computation~Self-organization}                       

\keywords{
    Oblivious Swarm Robots,
    Distributed Algorithm,
    Symmetries,
    Equivariant Dynamical Systems,
    Oblivious
}

\acknowledgements{}

\nolinenumbers 

\usepackage{tikz}
\usetikzlibrary{arrows}
\usetikzlibrary{shapes}

\usepackage{tikz-cd} 
\tikzcdset{scale cd/.style={every label/.append style={scale=#1},
    cells={nodes={scale=#1}}}}

\newcommand{\N}{\mathbb{N}}                                     
\newcommand{\R}{\mathbb{R}}                                     
\newcommand{\C}{\mathbb{C}}                                     
\providecommand{\abs}[1]{\left\lvert #1 \right\rvert}           
\providecommand{\norm}[1]{\left\lVert #1 \right\rVert}          

\newcommand{\Z}{\mathbb{Z}}

\def\oblot/{$\mathcal{OBLOT}$}
\def\LCM/{\textsc{LCM}}
\def\Look/{\textsc{Look}}
\def\Compute/{\textsc{Compute}}
\def\Move/{\textsc{Move}}

\newcommand{\fsync}{\textsc{$\mathcal{F}$sync}}
\newcommand{\gtm}{\textsc{Go-To-The-Middle}}
\newcommand{\GtC}{\textsc{Go-To-The-Center}}
\newcommand{\neargathering}{\textsc{Near-Gathering}}
\newcommand{\gathering}{\textsc{Gathering}}

\newcommand\set[1]{\left\{\,#1\,\right\}}

\DeclareMathOperator{\circulant}{circ}

\DeclareMathOperator{\Fix}{Fix}
\DeclareMathOperator{\Aut}{Aut}

\newcommand{\id}{\mathbf{I}}   
\newcommand{\eyes}{\mathbf{0}} 

\newcommand{\OO}{\mathbf{O}}

\renewcommand{\vec}[1]{\mathbf{#1}}

\newcommand{\z}{\vec{z}}
\renewcommand{\v}{\vec{v}}

\newcommand{\vzeta}{\boldsymbol{\zeta}}

\begin{document}

\maketitle

\begin{abstract}
In this article, we investigate symmetry properties of distributed systems of mobile robots.
We consider a swarm of $n\in \N$ robots in the \oblot/ model and analyze their collective \fsync\ dynamics using of equivariant dynamical systems theory.
To this end, we show that the corresponding evolution function commutes with rotational and reflective transformations of $\R^2$. These form a group that is isomorphic to $\mathbf{O}(2) \times S_n$, the product group of the orthogonal group and the permutation on $n$ elements.
The theory of equivariant dynamical systems is used to deduce a hierarchy along which symmetries of a robot swarm can potentially increase following an arbitrary protocol.
By decoupling the \Look/ phase from the \Compute/ and \Move/ phases in the mathematical description of an \LCM/ cycle, this hierarchy can be characterized in terms of automorphisms of connectivity graphs. 
In particular, we find all possible types of symmetry increase, if the decoupled \Compute/ and \Move/ phase is invertible. 
Finally, we apply our results to protocols which induce state-dependent linear dynamics, where the reduced system consisting of only the \Compute/ and \Move/ phase is linear.
\end{abstract}

\section{Introduction}\label{sec:intro}
Due to improved availability and decreased production costs of robotic units, the research area of coordination of \emph{swarms} of mobile, autonomous robots has seen great activity in recent years. Depending on their area of application, it can be economic or even the only possibility to employ large swarms of very simple and therefore cheap robots instead of highly specialized and capable units. A common theme in distributed computing is to overcome the lack of individual capabilities by clever coordination, that enable the swarm to collectively solve given tasks. This approach has seen great success \cite{H18,DBLP:series/lncs/FlocchiniPS19}. However, it comes with limitations or problems. In particular, when the swarm exhibits symmetries these can typically not be overcome by the collective. Here, we focus on mobile robots whose tasks consist of forming specific patterns based only on the observed positions of other robots. We apply the mathematical theory of dynamical systems to qualitatively investigate symmetries and their evolution.

\subparagraph{The \oblot/ Model}
We consider swarms of autonomous, mobile robots according to the the well-known \oblot/ model~\cite{DBLP:series/lncs/FlocchiniPS19}. Robots are assumed to be deterministic points that are \emph{oblivious} (no memory), \emph{anonymous} (no identities and indistinguishable), \emph{homogeneous} (execute the same algorithm), and \emph{disoriented} (no global coordinate system). We further assume robots to be \emph{silent} (no active communication).
Robots typically have a limited, constant \emph{viewing range}, within which they can perceive other robots. Identifying these positions is the only form of communication between robots.
The robots act in discrete time steps in which each of them performs a full LCM-cycle consisting of a \Look/- (observe surroundings), a \Compute/- (calculate target), and a \Move/- (move to target) phase.
These cycles are assumed to happen fully synchronously over all robots, which is referred to as the \emph{fully-synchronous} (\fsync) time model.
We consider only \emph{rigid} movements (robots always reach their designated target).
Powerful pattern formation algorithms have been developed in the \oblot/ model. The most general result to our knowledge states that any pattern can be formed by a swarm, if the initial condition is such that all robots are within the viewing range of every other robot (which can be realized by a distributed (\textsc{Near}-)\gathering{} algorithm in a previous phase) \cite{DBLP:conf/sand/HahnHK24}.
We would like to point out that numerous models with additional robot capabilities have been considered in the literature (e.g. \cite{DBLP:journals/siamcomp/CohenP05,DBLP:conf/wdag/IzumiKIW07,DBLP:series/lncs/LunaV19}) and the theory developed below can readily be adapted to many of them.

\subparagraph{Symmetries in Distributed Systems}
The problem with symmetries in this distributed setup can informally be summarized as -- \emph{robots that see the same thing, do the same thing}. More precisely, if two robots within their respective individual coordinate systems observe neighbors in exactly the same (relative) positions, they necessarily perform the same computations and move to the same point for the next round. As this is true for all robots, the symmetry of the swarm that relates the two robots remains intact in the next round and consequently for all future times. In the context of pattern formation by mobile robots, it has consequently been shown, that robot swarms can only form patterns which have the same or more (Euclidean) symmetries (as point clouds in $\R^2$), which is a direct consequence of these informal considerations. For further background, see~~\cite{DBLP:journals/siamcomp/SuzukiY99,DBLP:journals/siamcomp/FujinagaYOKY15,DBLP:journals/tcs/YamashitaS10,DBLP:series/lncs/Yamauchi19,DBLP:journals/dc/FlocchiniPSV17,DBLP:journals/siamcomp/SuzukiY06}.

The mathematical theory of \emph{equivariant dynamical systems} investigates the interplay between dynamics and symmetry properties and offers a wide set of tools for the formalization of symmetries of robot swarms and their temporal development. One of the main goals of this manuscript is to properly introduce this language and to set it up for further use in the study of arbitrary protocols in the \oblot/ model.
In \cite{GGHHK24}, this mathematical approach has been used to develop protocols which solve the \neargathering{} problem while not increasing the symmetries of a swarm.

\subparagraph{Our Contribution}
In this paper, we take a broader perspective. In particular, we describe the collective dynamics induced by an arbitrary distributed protocol in terms of a discrete dynamical system $\z^+=F(\z)$, where $\z$ is the collection of all robots' positions in an arbitrary global coordinate system. This system comes with natural symmetry properties, which relate to symmetries of the swarm. We investigate the interplay of both to unveil a hierarchical ordering by which symmetries might be gained in course of the evolution (cf. \Cref{thm:hierachy}). This hierarchy and the corresponding structural properties are subsequently further clarified by investigating only a single round and decoupling the \Look/ phase. This allows to describe the possible symmetries that can be gained in terms of automorphisms of the connectivity graph (cf. \Cref{prop:Gamma_G}). Finally, we make these considerations even more explicit in the case that the \Compute/ phase involves linear computations only.

\subparagraph{Outline}
The article is organized as follows. 
In~\Cref{sec:evolutionfunction}, we derive the mathematical description in terms of dynamical systems induced by arbitrary protocols according to the outlined robot model (\oblot/, \fsync, \LCM/) and analyze its equivariance properties. We further unveil the mechanisms by which the symmetries of a robot swarm change through the induced collective dynamics.
In~\Cref{sec:interpretation}, we interpret the symmetries (and their variations) in terms of the potential emerging behavior such as early collisions and invariant formations. We further deduce the hierarchy describing which symmetries can be gained by a given configuration.
By decoupling the \Look/ phase, we refine these considerations by effectively studying fixed interaction topologies. This is illustrated at the hand of a special class of protocols in~\Cref{subsec:LD}.
We conclude the article in~\Cref{sec:conclusion} with final remarks and an outlook.
Some technical proofs, examples, and details can be found in~\Cref{appendix}.

\section{Characterizing Protocols as Equivariant Dynamical Systems}%
\label{sec:evolutionfunction}
In this section, we mathematically describe a swarm protocol as a discrete dynamical system. We show that it is equivariant with respect to certain combinations of rotations, reflections, and permutations. Then, we formalize the notion of a symmetry of a swarm and provide first insights into the evolution of a swarm's symmetries under the collective dynamics. As most of these considerations have already been published in \cite{GGHHK24}, the presentation is streamlined. 

\subsection{Evolution Function}
\label{subsec:evolution_function}
We begin by recalling the three phases of a synchronous \LCM/ cycle. In the considered model, each robot synchronously
\begin{description}
	\item[Look:] observes the positions of other robots in relative coordinates within its vision range,
	\item[Compute:] computes a target point using the relative visible robots' positions, and
	\item[Move:] moves to the computed target point.
\end{description}
For our analysis, we take the stance of an external observer describing the dynamics in global coordinates. Given a swarm of $n\in \N$ robots we denote their positions in round $t$ by ${z_1^t=(x_1^t,y_1^t),\dotsc,z_n^t=(x_n^t,y_n^t) \in \R^2}$ in an arbitrary \emph{global coordinate system} in $\R^2$.
The (column) vector of all robot positions $\z^t= (z_1^t,\dotsc,z_n^t)^T \in(\R^2)^n\equiv\R^{2n}$ is the
\emph{configuration} of the entire swarm in round $t$.
During the analysis, we sometimes identify a robot with its position (e.g., saying that robot $z_i$ moves to its target point).

We now formalize the behavior of the swarm as a distributed system in the language of dynamical systems, which benefits the analysis in terms of the symmetry properties of the swarm. Recall that robots move according to synchronous \LCM/ cycles, which can most generally be described by a dynamical system of the form
\begin{equation*}
	z_i^{k+1} = f(z_i^k; \z^k) \quad \text{for } i=1,\dotsc, n,
\end{equation*}
for some suitable function $f\colon \R^2 \times \R^{2n} \to \R^2$. Here, the new position $z_i^{k+1}$ depends on its own as well as on (potentially) \emph{all other robots'} positions (e.g., if the robot currently sees the whole swarm and the protocol leads it to move towards the center of gravity of visible robots). As all robots perform the same protocol, the function $f$ does not depend on the index $i$. However, since all robots have their own coordinate system they must to be able to distinguish their own position from all other positions in the swarm.
This is reflected in the explicit first argument given to $f$.
The dynamics of the entire configuration is then described by the so-called \emph{evolution function} of the protocol $F:\R^{2n}\to\R^{2n}$ given by 
\begin{equation}
	\label{eq:general_configuration}
	\z^{k+1} = F(\z^k) = \begin{pmatrix} f(z_1^k; \z^k) \\ \vdots \\ f(z_n^k; \z^k)
	\end{pmatrix}.
\end{equation}
To shorten notation we may drop the explicit time superscript $t$ and abbreviate ${z_i^+ = f(z_i; \z)}$ where $z_i^+\in\R^2$ indicates the \enquote{next} position of robot $i$.
In the same way, ${\z^+ = F(\z)}$ denotes the \enquote{next} configuration under application of the evolution function $F$ to some configuration $\z \in \R^{2n}$. In mathematical terms, \eqref{eq:general_configuration} induces a \emph{discrete dynamical system}.

\subsection{Symmetries in Protocols}
\label{subsec:symmetries}
As pointed out in the introduction, symmetries play an important role when analyzing the collective behavior of a swarm. Most prominently, rotational and reflectional symmetries within the swarm in combination with permutations of robots turn out to lead to interesting dynamical phenomena.
To this end, we formalize the notion of a \emph{swarm symmetry} in such a way that we can apply the theory of \emph{equivariant} dynamical systems to analyze how a protocol's evolution function influences those symmetries. This generalizes the well-known notion of symmetricity \cite{DBLP:journals/siamcomp/FujinagaYOKY15} (cf. \Cref{rem:symmetricity} below).

\subparagraph{Equivariance of the Collective Dynamics}
Without loss of generality we assume the swarm to be centered at the origin of the global coordinate system. As the robots are unaware of the global coordinate system that we choose as the external observer, their computations and movements are insensitive to \emph{rotations} and \emph{reflections} of the global coordinates in the sense that collectively rotating or reflecting all robots' positions via a rotation or reflection $\rho\colon\R^2\to\R^2$ causes each robot to move to a rotated or reflected image of its original target point. Combined this implies that all \emph{orthogonal} symmetries $\mathbf{O}(2)$ are respected.
\begin{remark}
    \label{rem:chirality}
    Occasionally, it is assumed that all robots share a common sense of \emph{chirality}~\cite{DBLP:conf/sand/HahnHK24}. In this setting, robots are not insensitive to reflections, as these transformations reverse the chirality. Hence, only rotations $\mathbf{SO}(2)$ are respected. The arguments in the remainder of this article remain to hold in this restricted setup without any additional effort.
\end{remark}
On the other hand, recall that the robots are indistinguishable, which implies that, if a robot observes another robot in a certain position, it does not know which label this robot has. More precisely, the computations and movements of every single robot depend on the set of the other robots' positions rather than their ordered tuple. In other words, redistributing the labels via arbitrary permutations $\kappa\colon\set{1,\dotsc,n}\to\set{1,\dotsc,n}$ does not alter the computed target points.
In terms of the function governing the dynamics of all robots $f$, these observations can be reformulated as follows.
\begin{lemma}
	\label{lem:invariance}
	Let $\eta\in\R^2$ and $\vzeta=(\zeta_1,\dotsc,\zeta_n)^T\in\R^{2n}$ be arbitrary. The function governing the dynamics of all robots $f$ has the following properties:
	\begin{enumerate}[(i)]
		\item $f(\rho \eta; \rho \zeta_1, \dotsc, \rho \zeta_n) = \rho f(\eta; \mathbf{\zeta})$ for all rotations and refelections $\rho\colon\R^2\to\R^2$;
		\item $f(\eta; \zeta_{\kappa(1)}, \dotsc, \zeta_{\kappa(n)}) = f(\eta; \mathbf{\zeta})$ for all permutations $\kappa\colon\set{1,\dotsc,n}\to\set{1,\dotsc,n}$.
	\end{enumerate}
\end{lemma}

\begin{remark}
    For the sake of completeness, note that the computations and movements are also insensitive to \emph{(uniform) translations} of all robots' positions, i.e.,
    \begin{enumerate}[(iii)]
    \item $f(\eta + \xi; \zeta_1 + \xi, \dotsc, \zeta_n + \xi) = f(\eta; \mathbf{\zeta}) + \xi$ for all $\xi \in \R^2$;
    \end{enumerate}
    We typically neglect this property, as it has no practical implications in terms of \emph{invariant} configurations (cf. \Cref{def:symmetry}) in the analysis below.
\end{remark}
To shorten the statement in \Cref{lem:invariance} we lift the rotation $\rho$ and permutation $\kappa$ to the configuration space $(\R^2)^n \cong \R^{2n}$ using the following block matrices $M_{\rho}$ and $M_{\kappa}$ ($n \times n$ matrices with entries in $\R^{2 \times 2}$) generated by $\rho$ and $\kappa$:
\begin{align}%
\begin{split}
M_\rho = \id_n \otimes \rho &= 
\begin{pmatrix}
    \rho & \eyes & \cdots & \eyes \\
    \eyes & \rho & \ddots & \vdots \\
    \vdots & \ddots & \ddots & \eyes \\
    \eyes & \cdots & \eyes & \rho
\end{pmatrix}
\text{ and }(M_\kappa)_{ij} = \begin{cases}
    \id_2, \quad &\text{if } \kappa(i)=j\\
    \eyes, \quad &\text{otherwise}.
\end{cases}
\end{split}
\end{align}
Therein, $\id_l \in \R^{l\times l}$ denotes for the $l \times l$ identity matrix, $\eyes$ the zero matrix of suitable dimensions, and $\otimes$ the Kronecker product. In an abuse of notation, we identify both matrices with their $\R^{2n\times2n}$ counterparts.
This allows us to restate the properties of $f$ in \Cref{lem:invariance} for a configuration as follows.
\begin{enumerate}[(i)]
    \item $f(\rho \eta; M_\rho\mathbf{\zeta}) = \rho f(\eta; \mathbf{\zeta})$ for all rotations and reflections $\rho\colon\R^2\to\R^2$;
    \item $f(\eta; M_\kappa\mathbf{\zeta}) = f(\eta; \mathbf{\zeta})$ for all permutations $\kappa\colon\set{1,\dotsc,n}\to\set{1,\dotsc,n}$.
\end{enumerate}
On the global scale in regard of the entire configuration, the symmetry properties of $f$ imply that the evolution is insensitive to arbitrary rotations, reflections, and permutations of the robots in the following sense: It does not matter if first all robots perform a round of a \LCM/ cycle and then rotate/reflect/permute or do it the other way around. 
We define the set of all \emph{potential symmetries}
\begin{equation*}%
\Gamma = \set{\gamma = M_\kappa M_\rho \mid \kappa \text{ permutation}, \rho \text{ rotation or reflection}}.
\end{equation*}

\begin{theorem}[Protocols are equivariant]
	\label{thm:equiv}
	The evolution function $F$ is symmetric -- or \emph{equivariant} -- with respect to all potential symmetries of formations $M_\kappa M_\rho \in \Gamma$, i. e, 
	\begin{equation}
		\label{eq:equiv}
		F \circ (M_\kappa M_\rho) = (M_\kappa M_\rho)\circ F.
	\end{equation}
\end{theorem}
The proof of this statement can be found in \cite{GGHHK24} and for completeness also in \Cref{proof:equiv}.

\subparagraph{(Global) Symmetries of Robot Swarms}
Intuitively, a symmetry of a swarm's configuration should be a transformation of $\R^2$ which leaves the \emph{set} of robot positions unchanged. However, as symmetries of sets are somewhat in the mathematical dynamical analysis, extra care has to be taken to reformulate this idea in terms of tuples of positions. In particular, a transformation of $\R^2$ is a symmetry of a configuration $\z$ if it yields an arbitrarily permuted tuple of the same positions. In particular, we define
\begin{definition}[Symmetry of a Configuration]\label{def:symmetry}
	Consider a configuration $\z \in \R^{2n}$. Then $\gamma = M_\kappa M_\rho$ is a \emph{symmetry of the configuration $\z$} if and only if
	\begin{math}
	\gamma \z = {M}_\kappa {M}_\rho\z = \z
	\end{math}.
    The subset $\Gamma_\z \subseteq \Gamma$ of all such $\gamma \in \Gamma$ are called \emph{(actual) symmetries} of the configuration $\z$, i.e.,
    \begin{equation*}%
    \Gamma_\z = \set{\gamma = M_\kappa  M_\rho\in \Gamma \mid M_\kappa  M_\rho \z = \z}.
    \end{equation*}
\end{definition}
\begin{example}\quad\label{ex:triangle}
\begin{enumerate}[(a)]
    \item A configuration is fixed under all potential symmetries, i.e., $\Gamma_\z=\Gamma$, if and only if it is given by a swarm in which all robots have collided.
    \item\label{triangle_item}  Consider the robot swarm in \Cref{fig:triangle} in which three robots are arranged in an equilateral triangle. It is obvious that this swarm (i.e., the set of points $\set{z_1,z_2,z_3}$) is invariant under the symmetry group of the equilateral triangle $D_3$. However, the specifics of the formalization in \Cref{def:symmetry} are more subtle. In particular, the ordered tuple of robot position remains invariant only if after rotation the indices are shifted back and if after reflection two labels are interchanged. For a detailed discussion, see \Cref{eq:triangle_app} in the appendix.
    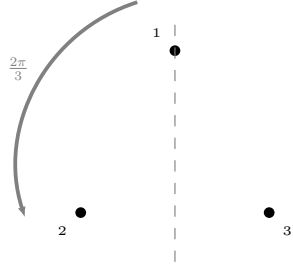
\begin{figure}
    \centering
    \resizebox{.3\linewidth}{!}{%
        \begin{tikzpicture}[
		xnode/.style = {
			circle,
			fill={black},
			inner sep=1pt
		},
		swarmx/.style = {
			circle,
			fill={black},
			opacity=1,
			inner sep=1pt
		},
		poly/.style = {
			regular polygon,
			regular polygon sides=3,
			inner sep=10,
			draw=none,
			thin,
			color=gray!20
		},
		cross/.style={
			cross out,
			draw, 
			minimum size=2pt, 
			inner sep=0pt,
			outer sep=0pt
		},
        scale=1.5
		]
		\node[poly] (p) {};
		\node[swarmx,label={[font=\tiny,scale=.7]135:$1$}] (1) at (p.corner 1) {};
		\node[swarmx,label={[font=\tiny,scale=.7]225:$2$}] (2) at (p.corner 2) {};
		\node[swarmx,label={[font=\tiny,scale=.7]315:$3$}] (3) at (p.corner 3) {};

        \node (a0) at (-.15,1) {};
		\node (a1) at (-.9,-0.45) {};
		\draw[-{Latex[length=1mm,width=.75mm]},line width=1pt,color=gray] (a0) to [bend right=45] node[midway, above left, scale=.5]{$\frac{2\pi}{3}$} (a1);
        
        \coordinate[above of=p, node distance=1.25cm] (b0);
        \coordinate[below of=p, node distance=1cm] (b1);    
        \draw[dashed, gray, line width=.3pt] (b0) to (b1);
	\end{tikzpicture}
    }
    \caption{A swarm configuration forming an equilateral triangle. The reflection axis of one reflection and rotation angle that both leave the swarm invariant are highlighted. For details see \Cref{ex:triangle}. (\ref{triangle_item})}
    \label{fig:triangle}
\end{figure}
\end{enumerate}
\end{example}

\begin{remark}
    \label{rem:symmetricity}
    A commonly used measure for ``how symmetric'' the configuration of a given robot swarm is is the notion of \emph{symmetricity}~\cite{DBLP:journals/siamcomp/FujinagaYOKY15}. It is defined by partitioning the robots into concentric regular polygons with the same number of sides (if possible). The largest possible number of sides for which such a partition exists is the symmetricity of the swarm. In other words, the symmetricity is the number of rotational symmetries of the swarm. Note that this number can also be deduced from our notion of symmetry by counting the number of different rotations $\rho$ whose corresponding maps $M_\rho$ are in $\Gamma_\z$, while ignoring the reflections (see also \Cref{rem:chirality}).
\end{remark}

\subparagraph{Algebraic Structure}
To make the mathematical description of the relevant symmetry transformations precise, we note that the sets of all potential symmetries $\Gamma$ and that of actual symmetries of a configuration $\Gamma_\z$ have underlying related algebraic structures:
\begin{definition}
    A \emph{group} is a set $G$ with an associative binary operation which contains an identity and inverses of each element. A \emph{subgroup} is a subset $H\subseteq G$ that is itself a group under the same binary operation, that is, it is closed under taking arbitrary products and powers and contains the inverses of all of its elements. A subgroup is typically denoted by $H\le G$. To abbreviate notation, one denotes by
    $\langle g_1, g_2, \dotsc \rangle \le G$
    the subgroup of $G$ that arises from taking arbitrary products, inverses and powers of the elements $g_1,g_2,\dotsc\in G$.
\end{definition}
\begin{proposition}
    The set of all potential symmetries forms a group with the binary operation given by matrix multiplication. In particular,
    $\Gamma \cong \OO(2)\times S_n$,
    where $\OO(2)$ is the \emph{orthogonal group on $\R^2$} and $S_n$ is the \emph{symmetric group on $n$ elements}.\footnote{The symbol ``$\cong$'' is read as ``is isomorphic to'' which means that there is a one-to-one map respecting the group structure.} The subset of actual symmetries of a configuration $\z\in\R^{2n}$ is a \emph{subgroup} $\Gamma_\z\le\Gamma$.
\end{proposition}
\begin{remark}
    To make the connection to equivariant dynamics more precise, we state.
    \begin{enumerate}[(a)]
        \item The map 
        $(\rho,\kappa)\mapsto M_\rho M_\kappa$
        is a \emph{faithful/injective representation} of $\OO(2)\times S_n$ on $\R^{2n}$.
        \item The subgroup of actual symmetries of a configuration $\Gamma_\z\subseteq \Gamma$ is called the \emph{isotropy subgroup} of the configuration $\z\in \R^{2n}$.
    \end{enumerate}
\end{remark}

\subparagraph{Evolution of Symmetries}
With this formalization at hand, we can study how the actual symmetries of a configuration evolve under the collective dynamics of the swarm induced by a given protocol. The 
mathematical framework for the investigation of dynamical systems with symmetry properties as in \Cref{thm:equiv} is given by so-called \emph{equivariant dynamical systems}, for which there exists a well developed theory to investigate the interplay of dynamics and symmetries (e.g. \cite{Golubitsky.1988,Chossat.2000}). For example, it can readily be shown that a protocol never causes the \emph{loss} of symmetries. In other words, symmetry cannot be broken. More precisely, we state the following result and refer to \Cref{proof:noloss} for its proof.

\begin{theorem}
    \label{thm:noloss}
	Consider the dynamics of a configuration $\z\in \R^{2n}$ according to an arbitrary protocol mathematically described as in \eqref{eq:general_configuration}. Then the configuration after one round cannot have fewer symmetries than the initial one, i. e, $\Gamma_{\z} \subseteq \Gamma_{\z^+}$.
\end{theorem}
In other words, symmetries cannot decrease for any protocol, but can potentially increase. In fact, this happens frequently in commonly investigated distributed protocols for example in achieving gathering via consecutive pairwise collisions (see for example the discussion of the \GtC{} protocol in \cite{DBLP:conf/opodis/CastenowH0KKH22}).
In the following section, we further investigate what symmetry increase means for a given swarm configuration.

Conversely, we also state that protocols which induce \emph{invertible} evolution functions \emph{preserve} symmetries \cite{GGHHK24}. Invertibility is to be understood on the level of the configuration in the eyes of the passive observer. 
This means that there exists a function $F^{-1}\colon\R^{2n}\to\R^{2n}$ such that $F( F^{-1}(\z))=F^{-1} (F(\z)) = \z$ for all $\z\in\R^{2n}$. Typically, $F^{-1}$ is not the evolution function of a distributed protocol.
In particular, a robot does not have to be able to determine its previous position (on its own).
The term ``locally'' further specifies that the inverse function needs to be defined only locally, meaning for all configurations that are sufficiently similar to a given one. For a more in-depth discussion, we refer to \cite{GGHHK24}.
\begin{proposition}\label{prop:preserve}
    If the evolution function $F$ is additionally (locally) invertible, then we also cannot \emph{gain} symmetries, i.e., $\Gamma_{\z^+} = \Gamma_{\z}$.
\end{proposition}

\section{A Closer Look on the Evolution of Symmetries}\label{sec:interpretation}
In this section, we take a more in-depth look at implications of the equivariance properties of the evolution function (\Cref{thm:equiv}) on the symmetries of a swarm (\Cref{def:symmetry}). We begin by describing immediate consequences of \Cref{thm:noloss}, i.e., of the fact that symmetries cannot be broken via collective dynamics. Then, we describe precisely how the symmetries of a swarm may increase if \Cref{prop:preserve} does not hold.
Everything discussed so far holds for arbitrary evolution functions $F$ and therefore for arbitrary protocols. To avoid uninteresting special cases---and the repetitive need to exclude those---we restrict ourselves to protocols which preserve connectivity. In particular, whenever two robots are within each other's perception range, they will remain visible to each other for all future times.

\subsection{Preservation of Structure}
\label{subsec:preservation}
\begin{figure}[!htb]
    \centering
    \subfloat[\centering Early collisions]{
        \label{fig:preservation01}
        \resizebox{.3\linewidth}{!}{%
            \begin{tikzpicture}[
    		swarmx/.style = {
    			circle,
    			fill={black},
    			opacity=1,
    			inner sep=1pt
    		},
            scale=.6
    		]
    		\node[swarmx,label={[font=\tiny,scale=.7]135:$1$}] (1) at (-1.5,0) {};
    		\node[swarmx,label={[font=\tiny,scale=.7,color=red]225:$2$}] (2) at (.7,.1) {};
    		\node[swarmx,label={[font=\tiny,scale=.7]135:$3$}] (3) at (-.75,-.5) {};
    		\node[swarmx,label={[font=\tiny,scale=.7]45:$4$}] (4) at (1.1,-.8) {};
    		\node[swarmx,label={[font=\tiny,scale=.7,color=red]135:$5$}] (5) at (.7,.1) {};
    		\node[swarmx,label={[font=\tiny,scale=.7]45:$6$}] (6) at (1.3,.4) {};
    		\node[swarmx,label={[font=\tiny,scale=.7]225:$7$}] (7) at (-.5,1.3) {};
    		\node[swarmx,label={[font=\tiny,scale=.7]315:$8$}] (8) at (-.4,-1.6) {};
    		
    	    \end{tikzpicture}
        }
    }\hfill
    \subfloat[\centering Regular polygons]{
        \label{fig:preservation02}
        \resizebox{.3\linewidth}{!}{%
            \begin{tikzpicture}[
    		swarmx/.style = {
    			circle,
    			fill={black},
    			opacity=1,
    			inner sep=1pt
    		},
    		poly/.style = {
    	   		regular polygon,
    			regular polygon sides=15,
    			inner sep=20,
    			draw,
    			thin,
    			color=gray!20
    		}
    		]
    		\begin{scope}[rotate=-30]
    			\node[poly] (p) {};
    			\node[swarmx] (1) at (p.corner 1) {};
    			\node[swarmx] (2) at (p.corner 2) {};
    			\node[swarmx] (3) at (p.corner 3) {};
    			\node[swarmx] (4) at (p.corner 4) {};
    			\node[swarmx] (5) at (p.corner 5) {};
    			\node[swarmx] (6) at (p.corner 6) {};
    			\node[swarmx] (7) at (p.corner 7) {};
    			\node[swarmx] (8) at (p.corner 8) {};
    			\node[swarmx] (9) at (p.corner 9) {};
    			\node[swarmx] (10) at (p.corner 10) {};
    			\node[swarmx] (11) at (p.corner 11) {};
    			\node[swarmx] (12) at (p.corner 12) {};
    			\node[swarmx] (13) at (p.corner 13) {};
    			\node[swarmx] (14) at (p.corner 14) {};
    			\node[swarmx] (15) at (p.corner 15) {};
    		\end{scope}        
    	   \end{tikzpicture}
        }
    }\hfill
    \subfloat[\centering Straight lines]{
        \label{fig:preservation03}
        \resizebox{.3\linewidth}{!}{%
            \begin{tikzpicture}[
    		swarmx/.style = {
    			circle,
    			fill={black},
    			opacity=1,
    			inner sep=1pt
    		}
    		]
    		\begin{scope}[rotate=-50]
                \coordinate (b0) at (0,1.1);
                \coordinate (b1) at (0,-1.2);\draw[gray, line width=.3pt] (b0) to (b1);
                
    			\node[swarmx] (1) at (0,0) {};
    			\node[swarmx] (2) at (0,1) {};
    			\node[swarmx] (3) at (0,-1.1) {};
    			\node[swarmx] (4) at (0,1.2) {};
    			\node[swarmx] (5) at (0,.5) {};
    			\node[swarmx] (6) at (0,-.6) {};
    			\node[swarmx] (7) at (0,-.25) {};
    			\node[swarmx] (8) at (0,.35) {};
    			\node[swarmx] (9) at (0,-1.2) {};
    			\node[swarmx] (10) at (0,.16) {};
    			\node[swarmx] (11) at (0,1.1) {};
    			\node[swarmx] (12) at (0,-.8) {};
    			\node[swarmx] (13) at (0,.9) {};
    			\node[swarmx] (14) at (0,.45) {};
    			\node[swarmx] (15) at (0,-1.35) {};
    		\end{scope}        
    	    \end{tikzpicture}
        }
    } \\
    \subfloat[\centering Reflection symmetries]{
        \label{fig:preservation04}
        \resizebox{.3\linewidth}{!}{%
            \begin{tikzpicture}[
    		swarmx/.style = {
    			circle,
    			fill={black},
    			opacity=1,
    			inner sep=1pt
    		}
    		]
    		\begin{scope}[rotate=-30]
    			\node (b0) at (0,-.5) {};
    			\node (b1) at (0,1) {};
    			\draw[gray!20, line width=.3pt] (b0) to (b1);
    			
    			\node[swarmx] (1) at (0,0) {};
    			\node[swarmx] (2) at (.5,.5) {};
    			\node[swarmx] (3) at (-.5,.5) {};
    			\node[swarmx] (4) at (1.1,-.2) {};
    			\node[swarmx] (5) at (-1.1,-.2) {};
    			\node[swarmx] (6) at (1.3,.4) {};
    			\node[swarmx] (7) at (-1.3,.4) {};
    		\end{scope}
    	    \end{tikzpicture}
        }
    }\hfill
    \subfloat[\centering Star shapes]{
        \label{fig:preservation05}
        \resizebox{.3\linewidth}{!}{%
            \begin{tikzpicture}[
    		swarmx/.style = {
    			circle,
    			fill={black},
    			opacity=1,
    			inner sep=1pt
    		},
    		poly/.style = {
    	   		regular polygon,
    			regular polygon sides=7,
    			draw,
    			thin,
    			color=gray!20
    		}
    		]
    		\node[poly, inner sep=20] (p1) {};
    		\node[swarmx] (1) at (p1.corner 1) {};
    		\node[swarmx] (2) at (p1.corner 2) {};
    		\node[swarmx] (3) at (p1.corner 3) {};
    		\node[swarmx] (4) at (p1.corner 4) {};
    		\node[swarmx] (5) at (p1.corner 5) {};
    		\node[swarmx] (6) at (p1.corner 6) {};
    		\node[swarmx] (7) at (p1.corner 7) {};
    	
    		\node[poly, inner sep=10, rotate=25.7] (p2) {};
    		\node[swarmx] (8) at (p2.corner 1) {};
    		\node[swarmx] (9) at (p2.corner 2) {};
    		\node[swarmx] (10) at (p2.corner 3) {};
    		\node[swarmx] (11) at (p2.corner 4) {};
    		\node[swarmx] (12) at (p2.corner 5) {};
    		\node[swarmx] (13) at (p2.corner 6) {};
    		\node[swarmx] (14) at (p2.corner 7) {};
    		
            \draw[red!60, line width=.6pt] (1) to (8) to (2) to (9) to (3) to (10) to (4) to (11) to (5) to (12) to (6) to (13) to (7) to (14) to (1);       
    	    \end{tikzpicture}
        }
    }\hfill
    \subfloat[\centering Concentric polygons]{
        \label{fig:preservation06}
        \resizebox{.3\linewidth}{!}{%
            \begin{tikzpicture}[
    		swarmx/.style = {
    			circle,
    			fill={black},
    			opacity=1,
    			inner sep=1pt
    		},
    		poly/.style = {
    			regular polygon,
    			regular polygon sides=5,
    			draw,
    			thin,
    			color=gray!20
    		}
    		]
    		\node[poly, inner sep=30, rotate=-30] (p1) {};
    		\node[swarmx] (1) at (p1.corner 1) {};
    		\node[swarmx] (2) at (p1.corner 2) {};
    		\node[swarmx] (3) at (p1.corner 3) {};
    		\node[swarmx] (4) at (p1.corner 4) {};
    		\node[swarmx] (5) at (p1.corner 5) {};
    		
    		\node[poly, inner sep=20, rotate=2] (p2) {};
    		\node[swarmx] (6) at (p2.corner 1) {};
    		\node[swarmx] (7) at (p2.corner 2) {};
    		\node[swarmx] (8) at (p2.corner 3) {};
    		\node[swarmx] (9) at (p2.corner 4) {};
    		\node[swarmx] (10) at (p2.corner 5) {};
    		
    		\node[poly, inner sep=15, rotate=23] (p3) {};
    		\node[swarmx] (11) at (p3.corner 1) {};
    		\node[swarmx] (12) at (p3.corner 2) {};
    		\node[swarmx] (13) at (p3.corner 3) {};
    		\node[swarmx] (14) at (p3.corner 4) {};
    		\node[swarmx] (15) at (p3.corner 5) {};
    		
    		\node[poly, inner sep=7, rotate=11] (p4) {};
    		\node[swarmx] (16) at (p4.corner 1) {};
    		\node[swarmx] (17) at (p4.corner 2) {};
    		\node[swarmx] (18) at (p4.corner 3) {};
    		\node[swarmx] (19) at (p4.corner 4) {};
    		\node[swarmx] (20) at (p4.corner 5) {};
    		
    		\node[poly, inner sep=3, rotate=11] (p5) {};
    		\node[swarmx] (21) at (p5.corner 1) {};
    		\node[swarmx] (22) at (p5.corner 2) {};
    		\node[swarmx] (23) at (p5.corner 3) {};
    		\node[swarmx] (24) at (p5.corner 4) {};
    		\node[swarmx] (25) at (p5.corner 5) {};
    	    \end{tikzpicture}
        }
    }
    \caption{Examples of structural properties that cannot be broken or resolved by the collective dynamics induced by any protocol according to the \oblot/ model. Labels and lines are only included to improve clarity.}
    \label{fig:preservation}
\end{figure}
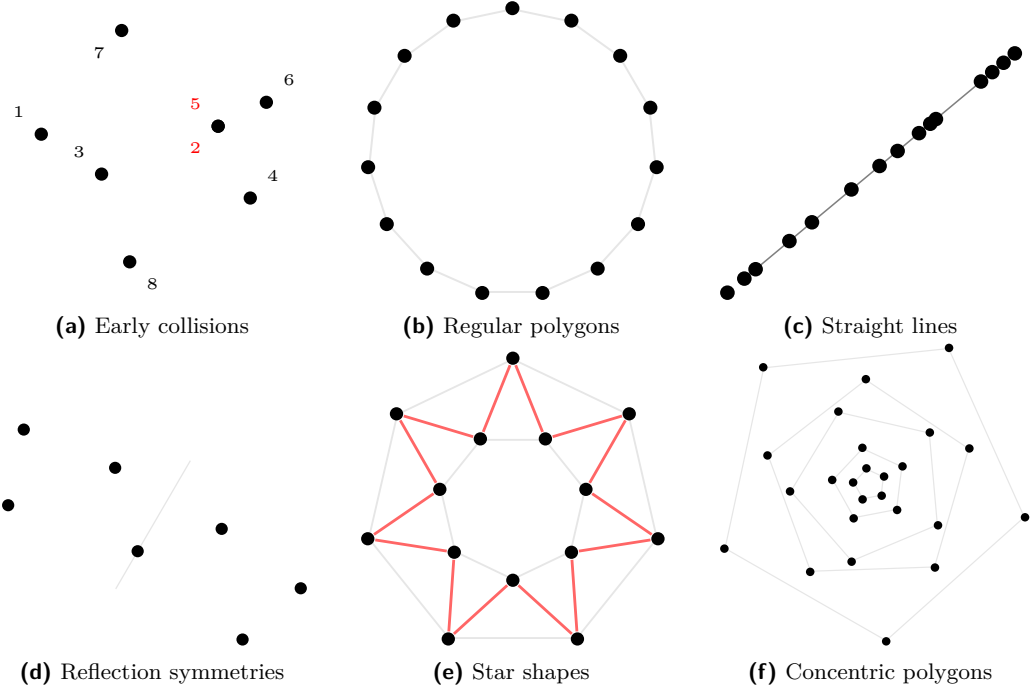

It is well-known that certain structural properties of a swarm do not disappear in its temporal evolution. These include symmetries but can occasionally be interpreted in terms of geometrical shapes which remain intact (while scaling and rotating). To make more sense of this general statement consider the following examples which are illustrated in \Cref{fig:preservation}.
\begin{description}
    \item[Early collisions:] Once two (or more) robots have collided, they cannot resolve this situation and move to different points in subsequent rounds.
    \item[Regular polygons:] When a swarm is arranged in the shape of a regular $m$-gon, it will remain in the form of a regular $m$-gon. If $m\lneq n$ this arrangement requires early collisions.
    \item[Straight lines] A collinear swarm cannot break its collinearity.
    \item[Reflection symmetries:] Similarly, reflection symmetries cannot be broken in the most general \oblot/ model (cf. \Cref{rem:chirality}).
    \item[Star shapes] If the swarm can be partitioned into two concentric $m$-gons, the formation can be interpreted as a star. This shape will remain but can degenerate into two concentric $m$-gons with the same radius (cf. \Cref{subsec:increase} below).
    \item[Arbitrary combinations of regular polygons:] The same holds true for finitely many regular $m$-gons, which may only degenerate into $m$-gons with the same radius.
\end{description}
Some of these preserved structural properties are obvious or at least easy to be deduced via combinatorial geometric considerations. Others, however, might be more surprising. Interestingly, all of these observations can be obtained as immediate consequences of the mathematical theory of equivariant dynamical systems outlined in \Cref{sec:evolutionfunction}.
In particular, for any subgroup $H\le\Gamma$ the set
\[
    \Fix(H)=\set{\z \in \R^{2n} \mid \gamma \z=\z \text{ for all } \gamma\in H},
\]
is a subspace of the configuration space which is called the \emph{fixed point subspace} of $H$. It is a standard result in equivariant dynamics, that any such fixed point subspace is invariant under the dynamics induced by $F$, i.e.
\begin{equation}
    \label{eq:fixed_invariant}
    F(\Fix(H)) \subseteq \Fix(H).
\end{equation}
For a proof we refer to \Cref{proof:fixed_invariant}. This means, that any configuration in one of these fixed point subspaces can only evolve into a configuration that is again in that same fixed point subspace. Naturally, any configuration is in the fixed point subspace of its own isotropy subgroup
$\z\in\Fix(\Gamma_\z)$,
In fact, the dynamical invariance of fixed point subspaces is equivalent to the statement in~\Cref{thm:noloss}.

Each of the preserved structural properties listed above precisely relates to a fixed point subspace of a suitable subgroup of transformations of the configuration space:
\begin{description}
    \item[Early collisions:] A configuration in which two robots are in the same position is fixed under $M_\kappa$, where $\kappa$ is the transposition of the corresponding labels. When multiple robots are collided, the configuration is invariant under arbitrary permutations of their labels.
    \item[Regular polygons:] A configuration arranged in a regular $m$-gon is fixed under $M_\kappa M_\rho$, where $\rho$ is the rotation by $\frac{2\pi}{m}$ and $\kappa$ suitably rearranges the labels (e.g. shifts them by one if the robots were labeled consecutively along the polygon). The isotropy subgroup emerges by forming arbitrary products (powers) of $M_\kappa M_\rho$
    \item[Straight lines:] A collinear swarm is fixed by $M_\rho$, where $\rho$ is the reflection through the line through the robots.
    \item[Reflection symmetries:] Similarly, a configuration with a reflection symmetry is fixed under $M_\kappa M_\rho$, where $\rho$ is the reflection and $\kappa$ suitably rearranges the labels.
    \item[Star shapes] A star shape has the same rotational symmetry as a polygon with vertices given by the outer vertices of the star. The argument then holds as above.
    \item[Arbitrary combinations of regular polygons:] The same holds true for finitely many concentric regular $m$-gons, which have the same rotational symmetry as one $m$-gon.
\end{description}
It is likely that more structural properties or shapes of a configuration can be identified in terms of symmetry properties. In the next subsection, we further elucidate the interaction of different patterns.

\begin{remark}
    Recall from \Cref{prop:preserve} that symmetries of a configuration cannot increase when the protocol is such that the evolution function is (locally) invertible. This can be interpreted similarly. In particular, a swarm cannot evolve into configuration that is an element in one of the fixed point subspaces if it was not in there initially. For instance, any collisions (including complete gathering) or the emergence of regular polygons is impossible in this situation. Instead, configurations can only converge towards such a structure.
\end{remark}

\subsection{Hierarchical Increase of Symmetries}
\label{subsec:increase}
In this section, we will make precise \emph{how} symmetries of a swarm may increase, if \Cref{prop:preserve} fails to hold, that is, if the evolution function $F$ is not (locally) invertible. This precise description holds for a general protocol whose evolution is not (locally) invertible. Therefore, the discussion demonstrates \emph{all possible} mechanisms by which symmetry may increase. A given protocol does not have to increase symmetries along those lines (or even at all for that matter, \Cref{prop:preserve} is a sufficient condition not a necessary one). First, we want to give some intuition at the hand of an example.
\begin{example}
    \label{ex:star_poly}
    Consider a swarm of $16$ robots arranged in the star shape represented by the black dots in~\Cref{fig:starfull}. Robot $i$ has precisely two visible neighbors $i-1$ and $i+1$ (counted modulo $16$). In each round the robots move $\approx0.52$ times the distance towards the midpoint of their observed neighbors. After one round, the swarm is in the configuration indicated by the red dots, which forms a regular $16$-gon. In particular, this configuration has $16$ rotational symmetries, while the prior star shape only had $8$ rotational symmetries.
    This specific example will be discussed in more detail in~\Cref{subsec:LD} below.
\end{example}
The observed symmetry increase in the previous example can again be seen as one instance of a general phenomenon in equivariant dynamics. In fact, due to the dynamical invariance of fixed point subspaces, it can readily be seen that the symmetries of a swarm can \emph{only} change, if the ``next'' configuration is inside a nested sub-fixed point subspace:
$\Fix(\Gamma_{\z^+}) \subsetneq \Fix(\Gamma_{\z^+})$.
In this case, the isotropy subgroup of $\z^+$ is a supergroup of the isotropy subgroup of $\z$:
$\Gamma_{\z^+} \gneq \Gamma_\z$.
More precisely, it is well known, that the isotropy subgroups form a lattice~\cite{Golubitsky.1988} and the same is true for corresponding fixed point subspaces (in which the inclusion relation holds in the opposite direction):
\begin{equation}
    \label{eq:lattice}
    \begin{tikzcd}[scale cd=0.53,every arrow/.append style={dash}]
        &&\Gamma\ar[d]\ar[dl]\ar[dr]\ar[drr]\\
        &\Gamma_\z \ar[d]\ar[dr]    &\Gamma_{\z'}\ar[d] &\Gamma_{\z''}\ar[d]\ar[dl]&\Gamma_{\z'''}\ar[dl]\ar[d]\\
        &\vdots\ar[d]    &\vdots\ar[d]\ar[rd]\ar[ld]   &\vdots\ar[d]    &\vdots\ar[ld]\ar[d]\\
        &\Gamma_{\tilde\z}\ar[dr]    &\Gamma_{\tilde\z'}\ar[d]   &\Gamma_{\tilde\z''}\ar[dl]    &\Gamma_{\tilde\z'''}\ar[lld]\\
        &&\set{\id_{2n}}
    \end{tikzcd}
    \begin{tikzcd}[scale cd=0.53,every arrow/.append style={dash}]
        &&\set{0}\ar[d]\ar[dl]\ar[dr]\ar[drr]\\
        &\Fix\left(\Gamma_\z\right) \ar[d]\ar[dr]    &\Fix\left(\Gamma_{\z'}\right)\ar[d] &\Fix\left(\Gamma_{\z''}\right)\ar[d]\ar[dl]&\Fix\left(\Gamma_{\z'''}\right)\ar[dl]\ar[d]\\
        &\vdots\ar[d]    &\vdots\ar[d]\ar[rd]\ar[ld]   &\vdots\ar[d]    &\vdots\ar[ld]\ar[d]\\
        &\Fix\left(\Gamma_{\tilde\z}\right)\ar[dr]    &\Fix\left(\Gamma_{\tilde\z'}\right)\ar[d]   &\Fix\left(\Gamma_{\tilde\z''}\right)\ar[dl]    &\Fix\left(\Gamma_{\tilde\z'''}\right)\ar[lld]\\
        &&\R^{2n}
    \end{tikzcd}
\end{equation}
This representation is highly stylized by representing the isotropy subgroups of arbitrary configurations.
The interplay of both lattices is to be understood as follows: A configuration $\z$, which does not have any non-trivial symmetries, is only left invariant by the identity, that is $\Fix(\Gamma_\z)=\set{\id_{2n}}$. The corresponding fixed point space is the full configuration space $\R^{2n}$, as the identity leaves all configurations unchanged. Both, isotropy subgroup and fixed point subspace, are at the lowest point in their respective lattice. On the other hand, a fully symmetric configuration $\z'$ satisfies $\Gamma_{\z'}=\Gamma$. The only configuration that is left invariant under all potential symmetries is the origin in configuration space, that is when all robots are in the same position (which we have chosen to be the origin of the coordinate system in $\R^2$). This isotropy group and fixed point subspace are at the top of their respective lattices.

As discussed above, it follows from the dynamical invariance of fixed point subspaces that changing symmetries through collective dynamics can only occur by evolving into a configuration in a sub-fixed point subspace: That is, by moving upwards in the lattice of fixed point subspaces~\eqref{eq:lattice}. If this happens, it comes with an increase in symmetries of the configuration into a supergroup, that is, the symmetries increase in upwards direction in the lattice of isotropy subgroups~\eqref{eq:lattice} along the same connections as in the lattice of fixed point subspaces (intermediate steps can be skipped).
In that sense, there is a natural hierarchy along which symmetries can be increased. This discussion has shown the following.
\begin{proposition}
    \label{prop:increase}
    \label{thm:hierachy}
    Collective dynamics induced by any protocol according to the considered model\footnote{\oblot/, \fsync{}, \LCM/} can only increase symmetries of the swarm following the lattice of isotropy subgroups of $\Gamma$~\eqref{eq:lattice} upwards.
\end{proposition}
\begin{example}
    Consider the situation as in \Cref{fig:preservation05}. That is assume $n=2m$ and the robots are arranged in two concentric regular $m$-gons (which we may interpret as a star shape). This configuration is fixed under rotations by $\frac{2\pi}{m}$ (and any multiples thereof) with a suitable rearrangement of the labels. All configurations with this rotational symmetry form a subspace of configuration space. The configurations with a rotational symmetry by half that angle $\frac{\pi}{m}$ form a subspace. Hence, according to \Cref{prop:increase} the evolution into a $\frac{\pi}{m}$-rotationally symmetric configuration is possible. This is precisely the case if the two regular $m$-gons evolve into having the same radius and forming a single regular $2m$-gon as in \Cref{fig:preservation02} (compare to \Cref{ex:star_poly}).

    On the other hand, the collision of precisely two robots in the star does not yield a subspace of the $\frac{2\pi}{m}$-rotationally invariant configurations. Hence, such a collision is impossible due to \Cref{prop:increase}. Note that early collisions are not impossible in this situation. For example, if all robots of one of the $m$-gons collide at once, the symmetry increases admissibly. An explicit example for this situation can be found in~\Cref{ex:gtm}.
\end{example}

Further theoretical background can be found in~\Cref{app:lattice}. In~\Cref{ex:increase} in the appendix, we explicitly derive the symmetry lattice for a swarm consisting of $n=6$ robots.

\section{Temporary Fixed Interaction Topology -- A One-Step Analysis}\label{sec:application}
In this section, we further investigate possible symmetry gains in view of \Cref{sec:interpretation} that can occur in \emph{one step} of the underlying protocol \eqref{eq:general_configuration}.
To this end, we decouple the \Look/ phase from the \Compute/ and \Move/ phases in the \LCM/ cycle.
More precisely, we consider an arbitrary but fixed initial configuration $\z\in \R^{2n}$ of the swarm and assume that in the \Look/ phase a \emph{connectivity graph} $G=(\set{1,\dotsc,n},E)$ is constructed.
The vertices are given by the set of robots and $e=(i,j) \in E$ if and only if the robot $i$ and $j$ perceive each other. By the properties of the \oblot/ model, this interaction is \emph{symmetric} in the sense that $(i,j) \in E \iff (j,i) \in E$.
We assume that the construction of $G$ is invariant under rotations and reflections, meaning the same graph is generated for the initial configuration $\z\in \R^{2n}$ as for the transformed configuration $M_\rho \z\in \R^{2n}$ for any rotation/reflection.
For instance, this is the case if $e=(i,j) \in E$ if and only if $\norm{z_i-z_j}_2 \leq \mathcal{C}$ for a predefined fixed viewing range $\mathcal{C} > 0$, which is typically considered.
We emphasize that the complete connectivity graph $G$ is only available for the global external observer, whereas each robot only has local information on its neighbors.

The remaining \Compute/ and \Look/ steps are mathematically described by a reduced evolution  function $F_G:\R^{2n}\to\R^{2n}$, i.e., the update of $\z\in \R^{2n}$ is as follows
\begin{align}\label{eq:FG}
    \z^+ = F_G(\z).
\end{align}
By removing \Look/ from \LCM/ the interaction structure is temporarily fixed, which decreases the number of symmetries of the system, as, in general, the function computing the target point of each robot $f$ is not invariant under \emph{every} permutation $\kappa:\set{1,\dotsc,n}\to\set{1,\dotsc,n}$ anymore (cf. \Cref{lem:invariance}).
Rather, the reduced system is only invariant with respect to permutations $\kappa_G:\set{1,\dotsc,n}\to\set{1,\dotsc,n}$ that respect the graph $G$, which are called \emph{graph automorphisms}
\begin{align*}
    \Aut(G) = \set{\kappa_G \in S_n\mid (i,j) \in \iff (\kappa(i),\kappa(j)) \in E},
\end{align*}
i.e., one-to-one permutations that preserve the edge-vertex connectivity.
Combined with the remaining $\OO(2)$ symmetry, the reduced system~\eqref{eq:FG} is only equivariant with respect to 
\begin{align*}
    \Gamma(G) = \set{\gamma = M_\kappa M_\rho \mid \kappa \in \Aut(G), \rho \text{ rotation or reflection}} \cong \OO(2) \times \Aut(G).
\end{align*}
It follows that the previous results of \Cref{sec:evolutionfunction,sec:interpretation} hold for the reduced group $\Gamma(G)$.
In particular, if $F_G$ is invertible symmetries in $\Gamma(G)$ cannot
be gained according to \Cref{prop:preserve}.
However, additional symmetries $\gamma \in \Gamma = \OO(2) \times S_n$ that are \emph{not} in the subgroup $\Gamma(G)$ can still occur, even if $F_G$ is invertible, as the reduced system is \emph{not} equivariant with respect to the entire group $\Gamma$.
We state this important observation in the following proposition.
\begin{proposition}\label{prop:Gamma_G}
    Consider the reduced system $F_G$ in \eqref{eq:FG} with graph $G$ and initial configuration $\z\in \R^{2n}$. Assume that the full evolution function $F$ is not invertible. Then
    \begin{enumerate}[i]
    \item Symmetry gain from $\Gamma\setminus\Gamma(G) = \OO(2)\times \left(S_n\setminus \Aut(G)\right)$ is (always) possible.
    \item If $F_G$ is not invertible, symmetry gain from $\Gamma(G) = \OO(2) \times \Aut(G)$ is possible.
    \item If $F_G$ is invertible, $\Gamma\setminus \Gamma(G) = \OO(2)\times \left(S_n\setminus \Aut(G)\right)$ is the only set of symmetries that can be added. 
    That is, if $\gamma \in \Gamma_{z^+}$ and $\gamma \notin \Gamma_z(G)$ then $\gamma \in \Gamma\setminus \Gamma(G) = \OO(2)\times \left(S_n\setminus \Aut(G)\right)$.
\end{enumerate}
\end{proposition} 
In particular, \Cref{prop:Gamma_G} provides a more in-depth understanding of the lattice structure of isotropy subgroups discussed in~\Cref{thm:hierachy}. Heuristically speaking, if $F_G$ is invertible, novel symmetries can only emerge involving permutations that were not automorphisms of the connectivity graph. For example, collisions are only possible of robots, that were not the image of each other under any automorphism.

\begin{remark}
    The reasoning behind decoupling \Look/ from \Compute/ and \Move/ is that it allows for a fine grained analysis of the symmetries.
    In general, the evolution function $F$ in \eqref{eq:general_configuration} is \emph{not} invertible, which allows for a large set of possible additional symmetries.
    However, the reduced evolution function $F_G$ \emph{can be} invertible for some initial configurations $\z\in \R^{2n}$, respectively connectivity graphs $G$.
    Thus, the set of possible additional symmetries is easier to analyze because symmetry gain from $\Gamma(G)$ is not possible (cf. \Cref{prop:Gamma_G} iii).
    Moreover, checking for invertibility is generally simpler for the reduced system \eqref{eq:FG} since, for instance, it can be linear in special cases which is studied in the upcoming \Cref{subsec:LD}.
\end{remark}

\subsection{State-Dependent Laplacian Dynamics}\label{subsec:LD}
While in the previous section a general evolution function $F$, respectively reduced evolution function $F_G$, is considered, we now apply the results to an explicit \emph{class} of protocols called \emph{State-Dependent Laplacian Dynamics}.
Here, we assume that the connectivity graph $G$, which is constructed in the \Look/ phase, is weighted by a (configuration-dependent) \emph{weight matrix} $W(\z)=(w(\z)_{i,j})_{i,j=1}^{n}\in \R^{n\times n}$, which again by the properties of the \oblot/ model is \emph{symmetric}, i.e., $W=W^T$.
Note that this weight matrix $W\in \R^{n\times n}$ is only available for the external observer as is the graph $G$ itself.
In particular, the $i$-th row is computed by the $i$-th robot using only its local information. 
In the \Compute/ phase, the entries of this matrix $W\in \R^{n\times n}$ serve as coefficients for a \emph{linear} combination of neighboring robots to compute target points $y_i\in \R^2$.
Finally, in the \Move/ phase each robot moves an $h$-fraction to that computed position $y_i\in\R^2$, where $h \in [0,1]$ is a predefined (fixed) step size.
Since a negative weight corresponds to the neighboring position \emph{reflected at the origin} and would force a robot to intentionally \emph{move away} from that neighbor, which contradicts connectivity and/or (possible) \gathering{} properties, it is reasonable to restrict to non-negative weights $w_{i,j} \geq 0$.
All of these steps can be modeled by a system of the form
\begin{align}\label{eq:DS_W}
    \begin{aligned}
      z_i^+ &= (1-h) z_i + h \sum_{j=1}^n w(\z)_{i,j} z_j, \quad \text{resp. in collective coordinates \eqref{eq:general_configuration}},\\
      \z^+ &= (1-h)\z + h (W(\z)\otimes \id_2)\z  = \left[\left((1-h)\id_n + h W(\z)\right) \otimes \id_2 \right]\z.
    \end{aligned}
\end{align}
This is a discrete state-dependent linear dynamical system with state-dependent system matrix ${\mathbf{A}_h(\z) = \left((1-h)\id_n + h W(\z)\right) \otimes \id_2 \in \R^{2n\times 2n}}$. 

It is well known that the model \eqref{eq:DS_W} solves the \gathering{} problem, if and only if,  for every configuration $\z\in \R^{2n}$ the weight matrix $W(\z)\in \R^{n\times n}$ has a simple eigenvalue $\lambda_1=1$ with eigenvector $(1,\dotsc,1)^T\in\R^n$ and all other eigenvalues $\lambda_j\in \C$ satisfy $\abs{\lambda_j}<1$. In this case, each robot $i$ computes a \emph{convex} combination of its neighboring robots' positions, as the eigenvalue/-vector condition implies
\begin{align}\label{eq:consistent}
    \sum_{j=1}^n w_{i,j} = 1 \quad \text{for all } i=1,\dotsc,n.
\end{align}
Afterwards, it moves an $h$-fraction towards that target point.
In the following, we only consider gathering protocols, particularly 
\begin{align}\label{eq:gathering}
    \lambda_1=1 \text{ and } \abs{\lambda_j}<1 \text{ for all } j=2,\dotsc,n \text{ and } \z\in\R^{2n}.
\end{align}
Note that since $W = W^T$ by construction, the eigenvalues are in fact real, i.e., we may assume $\lambda_j \in [-1,1]$ from now on.

Following the argumentation in \Cref{sec:application} the reduced collective dynamics of \eqref{eq:DS_W} is 
\begin{equation}\label{eq:DS_W_fixed}
    \begin{split}
      \z^+ &= F_G(\z) = (1-h)\z + h (W\otimes \id_2)\z  = \left[\left((1-h)\id_n + h W\right) \otimes \id_2 \right]\z = \mathbf{A}_h\z,
    \end{split}
\end{equation}
which models the \Compute/ and \Look/ phases decoupled from the \Look/ phase as a linear system with (temporary) fixed system matrix $\mathbf{A}_h = \left[\left((1-h)\id_n + h W\right) \otimes \id_2 \right] \in \R^{2n\times 2n}$ (cf. \eqref{eq:FG}).
Again, note that this update is only valid for the current step with initial configuration $\z\in\R^{2n}$.

To apply the results derived in \Cref{sec:application} we have to check, when the dynamics described by \eqref{eq:DS_W_fixed} is (not) invertible. Fortunately, invertibility of the reduced system \eqref{eq:DS_W_fixed} is simply characterized by the spectrum of the system matrix $\mathbf{A}_h \in \R^{2n\times 2n}$ whose (doubled) eigenvalues $\sigma_{j,h}\in \C$ depending on $h\in [0,1]$, are computed as the spectral shift:
\begin{align}\label{eq:sigma}
    \sigma_{j,h} = 1-h + h\lambda_j = 1+h(- 1+\lambda_j) \quad \text{for } j=1,\dotsc,n,
\end{align}
Now, the dynamics in \eqref{eq:DS_W_fixed} is not invertible if and only there is a zero eigenvalue, i.e., 
\begin{align}\label{eq:sigma=0}
   \sigma_{j,h} = 0 \iff \lambda_j = \frac{h-1}{h}. 
\end{align}
Thus, by \Cref{prop:Gamma_G} symmetries in $\Gamma_G$ are preserved if $\lambda_j \neq \frac{h-1}{h}$ for all $j=1,\dotsc,n$. For instance, this is the case when $h < \frac{1}{2}$ since the protocol is assumed to be gathering (cf. \eqref{eq:gathering}). We summarize this first result in the following lemma,

\begin{lemma}\label{lem:h<0.5}
    A one-step reduced protocol of the form \eqref{eq:DS_W_fixed} with connectivity graph $G$ preserves its symmetries in $\OO(2)\times \Aut(G)$, if the step size is chosen small enough $h< \frac{1}{2}$.
\end{lemma}

Further analyzing the invertibility condition \eqref{eq:sigma=0} shows that only non-positive eigenvalues $\lambda_j \in [-1,0]$ can cause non-invertibility of the dynamics.
Indeed, by continuity in $h$ and the intermediate value theorem, in this case there is always a step size $h\in [0,1]$ such that the reduced system becomes non-invertible.
According to \Cref{lem:h<0.5} this step size is bounded from below by $h\geq \frac{1}{2}$.
We formalize this result in the following lemma. 

\begin{lemma}\label{lem:lambda_in_[-10]}
    Consider a one-step reduced protocol of the form \eqref{eq:DS_W_fixed} with weight matrix $W\in \R^{n\times n}$ and assume $\lambda_j\in [-1,0]$ is an eigenvalue of $W\in\R^{n\times n}$. Then, there is a step size $h\in [\frac{1}{2},1]$ such that the protocol is non-invertible in this step. In this case, \cref{thm:hierachy} is applicable and symmetry gain with respect to $\OO(2) \times \Aut(G)$ is possible.
\end{lemma}
To summarize, in order to guarantee invertibility for the decoupled \Compute/ and \Move/ system, it is sufficient to either assume $h<\frac{1}{2}$ or $\lambda_j\notin [-1,0]$ for $j=1,\dotsc,n$. For the latter case the step size $h\in [0,1]$ can be chosen arbitrary.  

In the linear case, one can answer the question, where do additional symmetries come from?
To this end, we consider the case where $F_G$ is not invertible, i.e., there is a zero eigenvalue $\sigma_{j,h} = 0$ with corresponding $k$-dimensional null space $\ker(\mathbf{A})\subseteq \R^{2n}$. In particular, the swarm state space $\R^{2n}$ can be decomposed into two subspaces such that
\begin{align}\label{eq:decomp_V0}
    \R^{2n} = \ker(\mathbf{A}) \oplus V \text{ for a suitable $n-k$-dimensional subspace $V\subseteq \R^{2n}$}.
\end{align}
This allows us to uniquely express the initial configuration $\z\in \R^{2n}$ as
\begin{align*}
    \z = \v_0 + \v \in \R^{2n}, \quad \text{where } \v_0 \in \ker(\mathbf{A}) \text{ and } \v \in V.
\end{align*}
Note that this decomposition is \emph{invariant} under the dynamics, particularly, $\mathbf{A}v \in V$.
Let $\gamma \in \Gamma_{\z^+}(G)$ be a symmetry of the updated configuration $\z^+\in \R^{2n}$ which is \emph{not} a symmetry of $\z\in \R^{2n}$ itself, i.e., it is an additional \emph{new} symmetry. Further assume $\gamma \in \Gamma(G)$ which is possible since $F_G$ is non-invertible (cf. \Cref{prop:Gamma_G}). By evolving $\v\in V$ under the dynamics we can compute
\begin{align*}
    \mathbf{A}\v = \mathbf{A}(\z - \v_0) = \mathbf{A}\z = \z^+ = \gamma \z^+ = \gamma \mathbf{A}\z = \gamma \mathbf{A} (\v + \v_0) = \gamma \mathbf{A}\v = \mathbf{A}\gamma \v, 
\end{align*}
where we used the symmetry properties of $\mathbf{A}\in \R^{2n\times 2n}$ and $\z\in \R^{2n}$. Now, $\v\in V \subseteq \R^{2n}$ does not contain any kernel vector of $\mathbf{A}\in \R^{2n\times 2n}$ by definition of the decomposition \eqref{eq:decomp_V0} and we conclude $\gamma \v = \v$.
Thus, the part $\v\in V$ which is \emph{not} in the kernel of $\mathbf{A}\in \R^{2n\times 2n}$ already has the new symmetry $\gamma\in \Gamma(G)$. 
In this sense, kernel vectors $\v_0\in \ker(\mathbf{A})$ are responsible for symmetry gain in $\Gamma(G)$ as they decrease the symmetry of $\v\in V$ by superposition. As they vanish under the dynamics \eqref{eq:DS_W_fixed} the ``actual'' symmetry of $\v\in V$ prevails for the entire configuration $\z^+\in \R^{2n}$.
This phenomenon is further illustrated in the following explicit example.

\begin{example}
    \label{ex:gtm}
    As a special case consider a \emph{circulant} interaction topology given by a \emph{circulant} graph $G_\text{circ}$ for which one may completely characterize the graph automorphisms
    \begin{align*}
    \Aut(G_\text{circ}) = \langle \kappa_1 \rangle := \set{\kappa^k \mid k \in \Z} = \set{e,\kappa_1,\kappa_1^2,\dotsc,\kappa_1^{n-1}} \cong C_n.
    \end{align*}
    Here, $\kappa_1\in S_n$ denotes the cyclic permutation shifting the indices by $1$ and $C_n$ is the \emph{cyclic group} on $n$ elements. In \cite{GGD25}, an extensive discussion on gathering speeds has been done.
    
    As indicated in \Cref{subsec:LD}, the eigenvalues of the weight matrix $W\in \R^{n\times n}$ are of particular importance for invertibility and symmetry investigations. Fortunately, all eigenvalues and eigenvectors can be analytically derived for a circulant matrix (see e.g. \cite{G05}) as
    \begin{align}\label{eq:EW}
    	\lambda_j = \sum_{i=0}^{n-1} w_i\omega^{ij} \in \R  \quad \text{for } j = 0,\dotsc,n-1,
    \end{align}
    where $w = (w_0,\dotsc,{n-1})^T \in \R^{n}$ is the \emph{generating} vector of the weight matrix $W\in \R^{n\times n}$ and $\omega = \exp{\left(\frac{2\pi \mathbf{i}}{n}\right)}$ is a primitive $N$-th root of unity. Here we shifted the (arbitrary) enumeration of the eigenvalues by $1$ as is makes the notation simpler.  Note that $\lambda_j \in \R$ since $W=W^T$.
    By \eqref{eq:consistent} the eigenvalue formula \eqref{eq:EW} can be seen as a \emph{convex} combination of roots of unity.
    The corresponding shifted eigenvalues $\sigma_{j,h} \in \R$ in \eqref{eq:sigma} of the system matrix are then
    \begin{align}\label{eq:sigma_circ}
       \sigma_{j,h} = 1-h + h \sum_{i=0}^{n-1} w_i\omega^{ij} \quad \text{for } j=0,\dotsc,n-1.
    \end{align}
    
    As an explicit protocol, we consider the \gtm{} protocol (cf. \cite{Kling.2011}). Its weight matrix is given by ${W = \circulant(0,\frac{1}{2},0,\dotsc,\frac{1}{2})}$, i.e., robot $i$ is influenced by its first neighbors to the left and the right.
    Using the eigenvalue formula \eqref{eq:sigma_circ} with ${w=(0,\frac{1}{2},0,\dotsc,\frac{1}{2})^T \in \R^n}$ we compute
    \begin{align*}
        \sigma_{j,h} = 1-h + h\cos\left(\frac{2\pi j}{n}\right) \quad \text{for }j = 0,\dotsc,n-1,
    \end{align*}
    which are convex combinations of $1$ and the cosine curve. The eigenvalues $ \sigma_{j,h} \in \R$ are visualized in \Cref{fig:EW_sigma} for different choices of $h\in [0,1]$.
    \begin{figure}[!htb]
        \centering
        \subfloat[\centering $n=15$]{
            \label{fig:sig_N15}
        	\includegraphics[width=.47\linewidth]{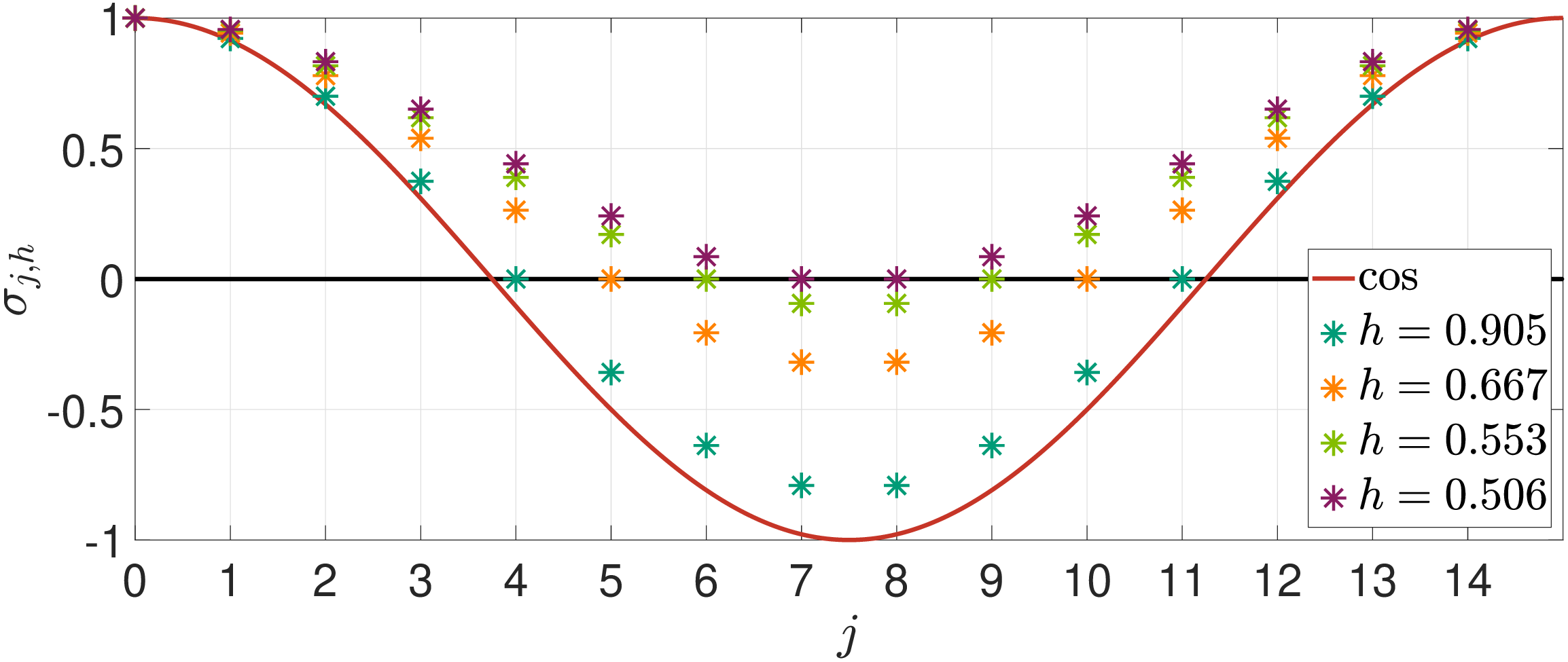}
        }
        \hfill
        \subfloat[\centering $n=16$]{
        	\label{fig:sig_N16}
        	\includegraphics[width=.47\linewidth]{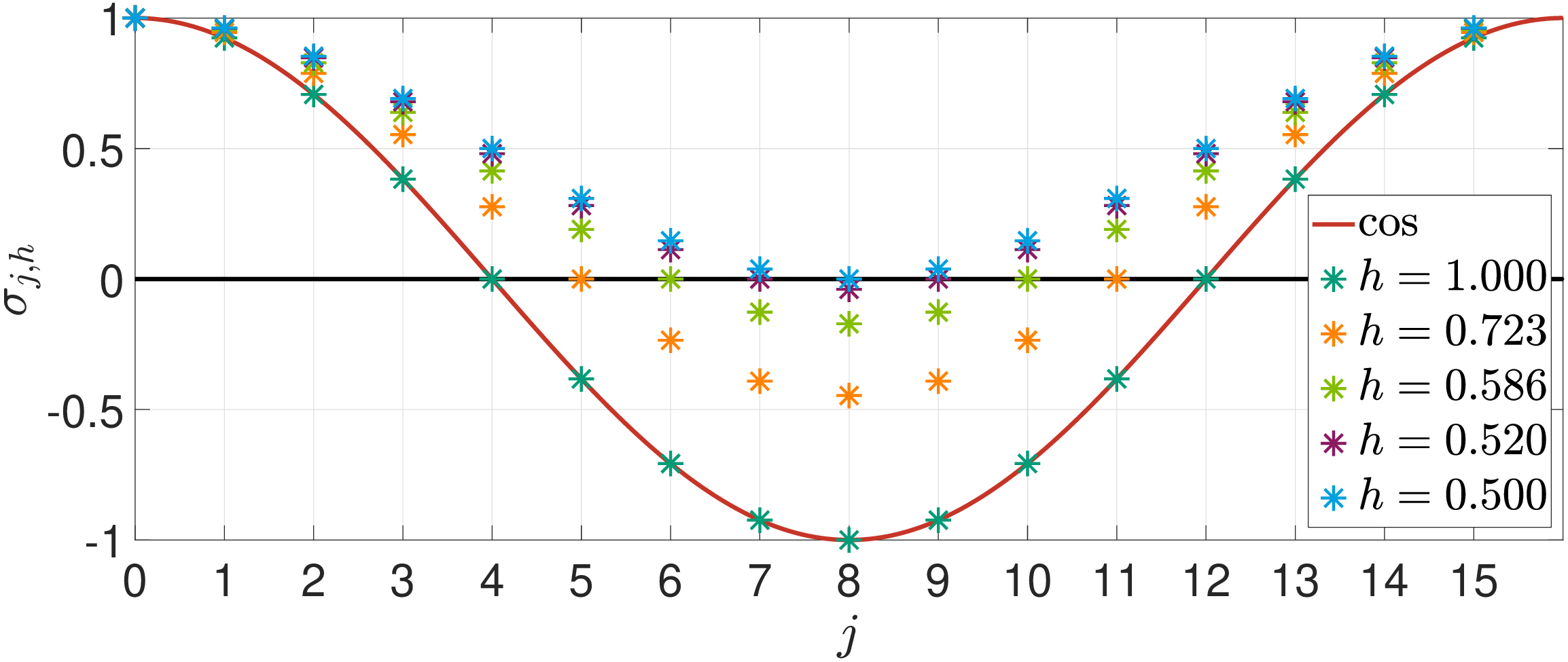}
        }
        \caption{Illustration of the eigenvalues $\sigma_{j,h} \in \R$ for different step sizes $h\in [0,1]$ such that there is a zero eigenvalue.}
        \label{fig:EW_sigma}
    \end{figure}
    In particular, we can \emph{increase} the symmetry by elements of $\Gamma(G_\text{circ})$ of configurations by choosing $h\in [0,1]$ appropriately. We visualize this phenomenon in \Cref{fig:star}.
    \begin{figure}[!htb]
    	\centering
        \subfloat[\centering $\z\in \R^{2n}$ and $\z^+ \in \R^{2n}$]{
            \label{fig:starfull}
            \includegraphics[width=.3\linewidth]{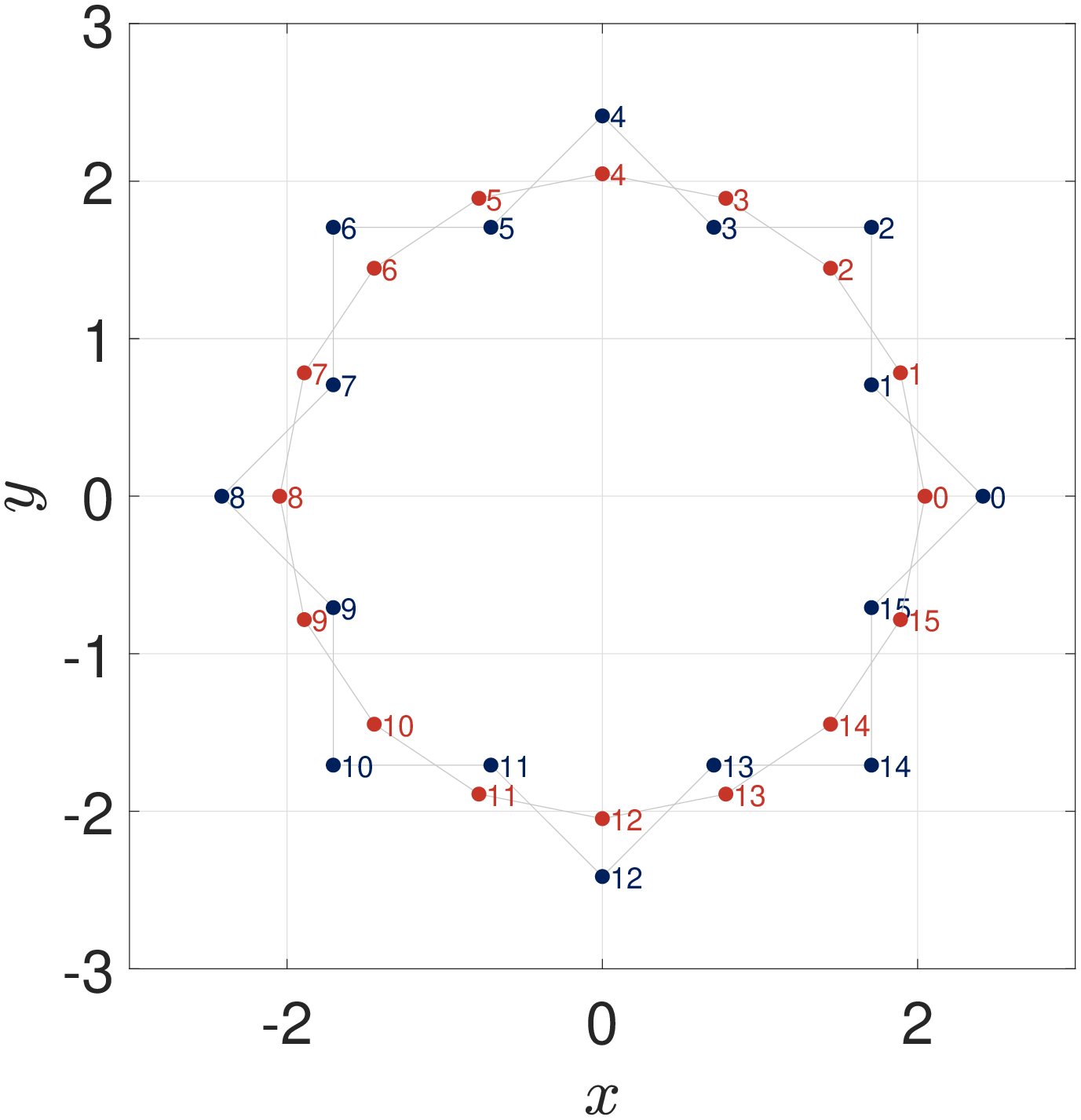}
        }
        \hfill
    	\subfloat[\centering $\v \in V$ and $\v^+ \in V$]{
    		\label{fig:vv+}
    		\includegraphics[width=.3\linewidth]{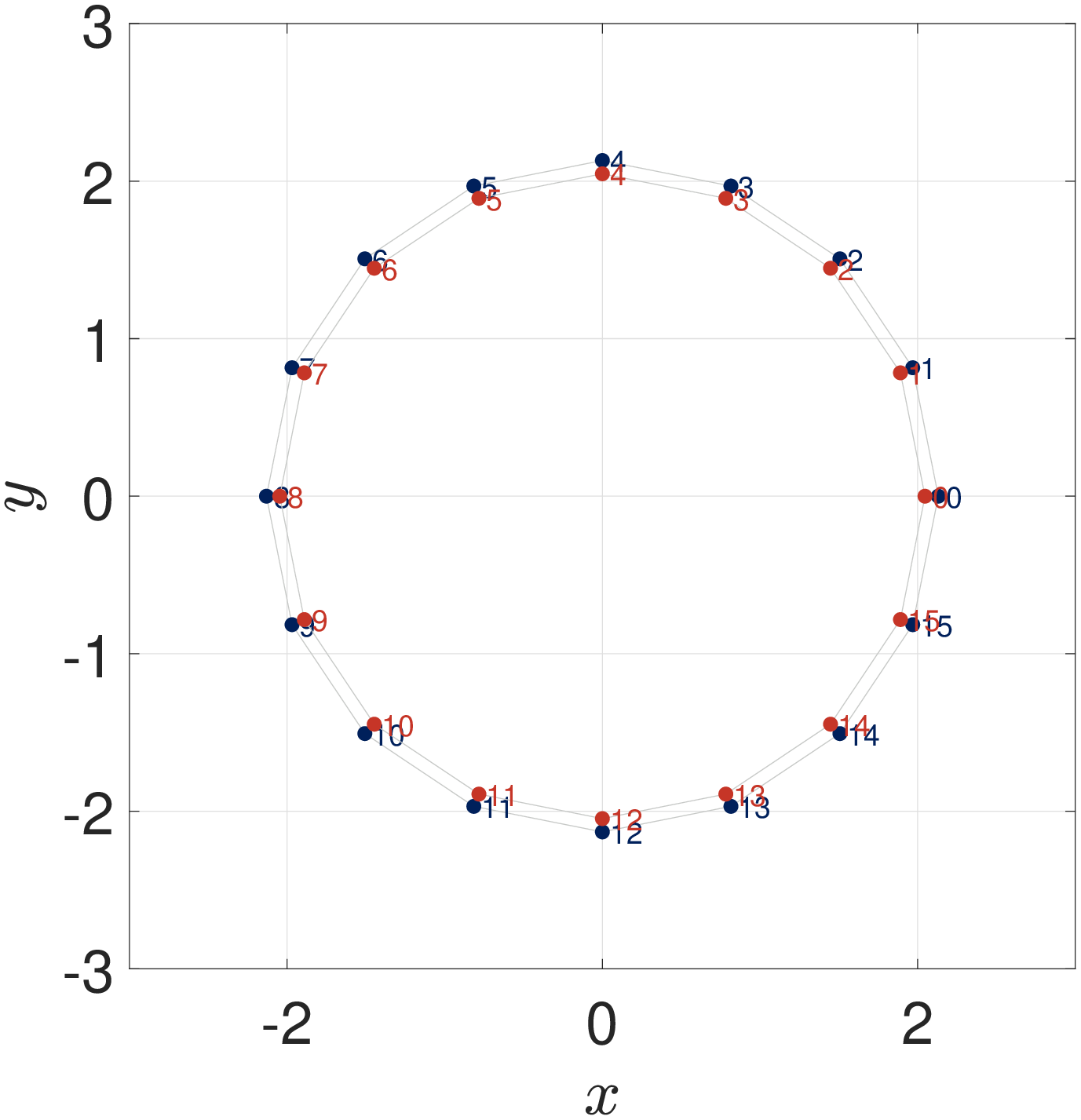}
    	}
        \hfill
    	\subfloat[\centering $\v_0 \in \ker(\mathbf{A})$]{
    		\label{fig:v0}
    		\includegraphics[width=.3\linewidth]{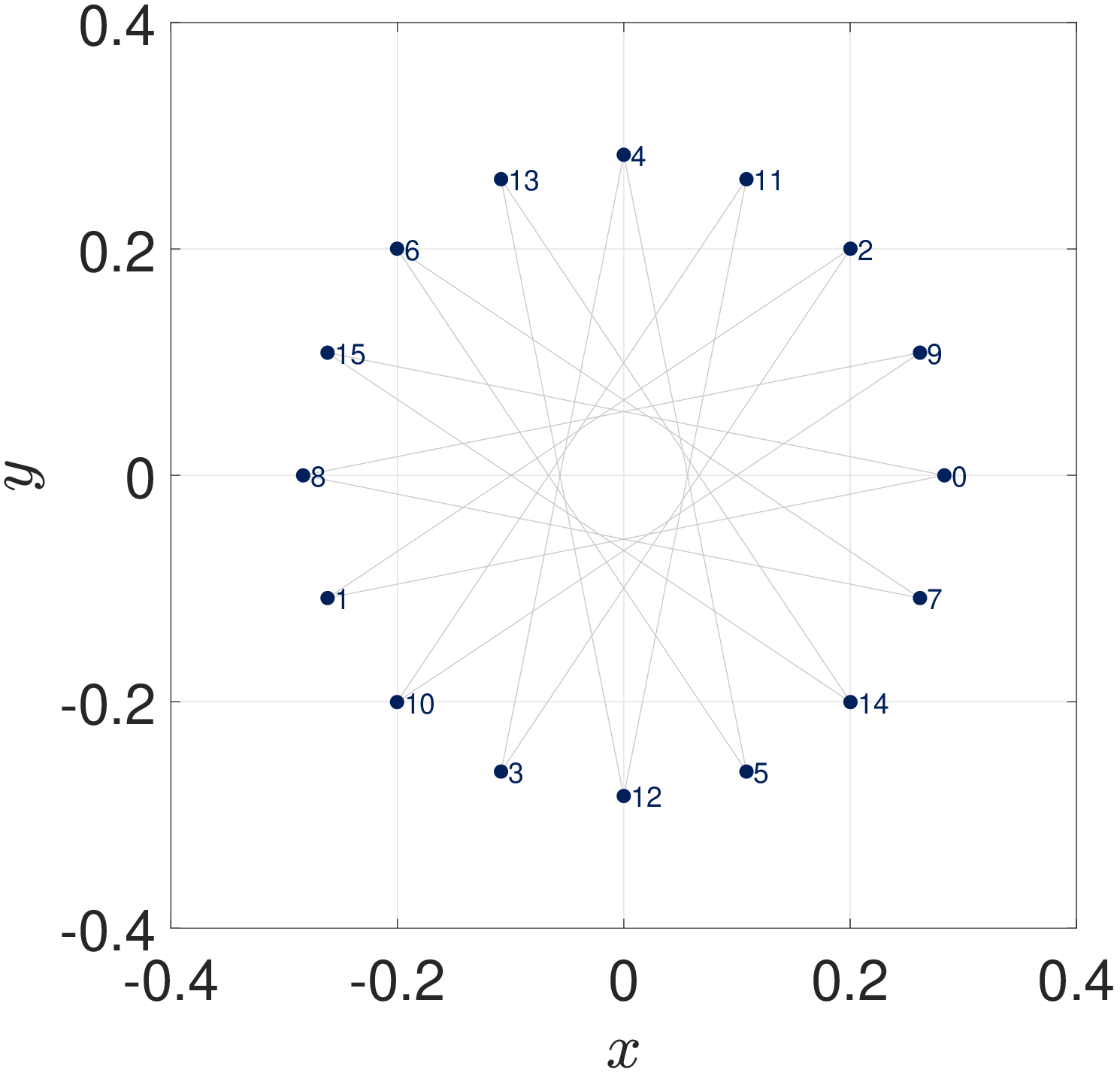}
    	}
        \caption{For $n=16$ a star shaped configuration $\z = \v_0 + \v \in \R^{2n}$, where $\v_0\in \ker(\mathbf{A})$ and $\v\in V$, is mapped to an $n$-gon shaped configuration $\z^+=F_G(\z)$ for $h\approx 0.520$ (cf. \Cref{fig:sig_N16}) increasing its symmetry under the dynamics \eqref{eq:DS_W_fixed}.}
    	\label{fig:star}
    \end{figure}
\end{example}

\section{Discussion \& Conclusion}\label{sec:conclusion}
In this article we investigated symmetry properties of arbitrary distributed swarm systems following the \fsync{} \oblot/ model and derived a hierarchy along which symmetries can emerge under iteration. In particular, they cannot disappear (cf. \Cref{thm:noloss}).
This hierarchy follows the lattice of isotropy groups and the corresponding fixed point spaces \eqref{eq:lattice}.

We exemplary illustrated our findings by decoupling the \Look/ phase from the \Move/ and \Compute/ phases in the \LCM/ cycle as invertiblity of the reduced system is presumably more likely to hold true (cf. \Cref{prop:preserve}), particularly, if the reduced system is linear.
In this situation, we showed that kernel vectors are responsible for additional new symmetries: They vanish under the dynamics and only the non-kernel part with those new symmetries remains.
Certainly, this is a special case and the analysis of more complex protocol remains for future research. We claim that for general non-linear protocols configurations that prevent the evolution function to be injective play the analogous as the kernel vectors.
In particular, we are convinced that that the presented approach provides a theoretical pathway for in-depth analysis of explicit non-linear protocols such as \GtC{} and could help with design of new distributed algorithms (e.g. for pattern formation \cite{DBLP:journals/siamcomp/SuzukiY99,DBLP:journals/tcs/YamashitaS10}).

Finally, in this work we considered the \fsync{} model. It remains an interesting open question if the results can be extended to \emph{asynchronous} time models \cite{DBLP:journals/siamcomp/FujinagaYOKY15,DBLP:series/lncs/Flocchini19}.

\bibliography{literature.bib}

\appendix
\section{Appendix}
\label{appendix}
\subsection{Proofs}\label{app:proofs}

\begin{proof}[Proof of \Cref{thm:equiv}]\label{proof:equiv}
    It suffices to prove that $F$ is equivariant as in \eqref{eq:equiv} with respect to $M_\kappa$ for all $\kappa$ and $M_\rho$ for all $\rho$ individually since in this case we have
	\begin{align*}
		F \circ (M_\kappa M_\rho) 	&= F \circ (M_\kappa \circ M_\rho)
									= (F \circ M_\kappa) \circ M_\rho 
									= (M_\kappa \circ F) \circ M_\rho \\
									&= M_\kappa \circ (F \circ M_\rho) 
									= M_\kappa \circ (M_\rho \circ F) 
									= (M_\kappa M_\rho) \circ F.
	\end{align*}
    Let $\vzeta=(\zeta_1,\dotsc,\zeta_n)^T\in\R^{2n}$ be a point in configuration space, ${\kappa\colon\set{1,\dotsc,n}\to\set{1,\dotsc,n}}$ a permutation, and $\rho\colon\R^2\to\R^2$ a rotation or reflection. Then, using \Cref{lem:invariance} we compute
	\begin{equation*}
		\begin{split}
			(F\circ M_\kappa)(\vzeta)	&= F(M_\kappa \vzeta)
										= \begin{pmatrix} f((M_\kappa \vzeta)_1, M_\kappa \vzeta) \\ \vdots \\ f((M_\kappa \vzeta)_n, M_\kappa \vzeta) \end{pmatrix}
										= \begin{pmatrix} f(\zeta_{\kappa(1)}, M_\kappa \vzeta) \\ \vdots \\ f(\zeta_{\kappa(n)}, M_\kappa \vzeta) \end{pmatrix}  \\
										&= \begin{pmatrix} f(\zeta_{\kappa(1)}, \vzeta) \\ \vdots \\ f(\zeta_{\kappa(n)}, \vzeta) \end{pmatrix}
										= \begin{pmatrix} F(\vzeta)_{\kappa(1)} \\ \vdots \\F(\vzeta)_{\kappa(n)} \end{pmatrix} 
										= M_\kappa F(\vzeta) 
										= (M_\kappa \circ F)(\vzeta),
		\end{split}
	\end{equation*}
	Similarly, we obtain
	\begin{equation*}
		\begin{split}
			(F\circ M_\rho)(\vzeta)	&= F(M_\rho \vzeta) 
									= \begin{pmatrix} f((M_\rho \vzeta)_1, M_\rho\vzeta) \\ \vdots \\ f((M_\rho \vzeta)_n, M_\rho\vzeta) \end{pmatrix} 
									= \begin{pmatrix} f(\rho \zeta_1, M_\rho\vzeta) \\ \vdots \\ f(\rho \zeta_n, M_\rho\vzeta) \end{pmatrix} \\
									&= \begin{pmatrix} \rho f(\zeta_1, \vzeta) \\ \vdots \\ \rho f(\zeta_n, \vzeta) \end{pmatrix} 
									= M_\rho F(\vzeta) 
									= (M_\rho F)(\vzeta).
		\end{split}
	\end{equation*}\end{proof}

\begin{proof}[Proof of \Cref{thm:noloss}]\label{proof:noloss}
	Let $\z\in\R^{2n}$ and $\gamma\in \Gamma_{\z}$. Then
	\begin{equation*}
		\gamma\z^+=\gamma F(\z)=F(\gamma\z)=F(\z)=\z^+,
	\end{equation*}
	which is due to the symmetry of $F$ by \Cref{thm:equiv} and fact that $\gamma$ leaves $\z$ unchanged.
\end{proof}

\begin{proof}[Proof of \eqref{eq:fixed_invariant}]\label{proof:fixed_invariant}
    Let $H\leq \Gamma$ be subgroup of $\Gamma$, $\z\in \Fix(H)$ and $\gamma\in H$. Then by \Cref{thm:equiv} we compute $\gamma F(\z) = F(\gamma\z) = F(\z)$, i.e., $F(\z) \in \Fix(H)$.
\end{proof}

\subsection{Lattice Structure}
\label{app:lattice}
In this appendix, we provide some additional details regarding the lattice structures of isotropy subgroups and fixed point subspaces laid out in \Cref{subsec:increase}. In general, a \emph{lattice} is a \emph{partially ordered set} together with a \emph{join operation} (to compute the least upper bound of two elements) and a \emph{meet operation} (to compute the greatest lower bound of two elements).

The set of isotropy subgroups is partially ordered with respect to the subgroup relation~$\le$. The join operation on this set is given by
\[ \Gamma_{\z} \vee_{\mathrm{iso}} \Gamma_{\z'} = \left\langle \Gamma_{\z} \cup \Gamma_{\z'} \right\rangle, \]
which is the subgroup of $\Gamma$ that emerges from taking arbitrary products of elements of both subgroups $\Gamma_{\z}$ and $\Gamma_{\z'}$. The meet operation in the lattice is given by intersection
\[ \Gamma_{\z} \wedge_{\mathrm{iso}} \Gamma_{\z'} = \Gamma_{\z} \cap \Gamma_{\z'}. \]

The set of fixed point subspaces, on the other hand, is partially ordered with respect to the subspace relation $\subseteq$. The join operation on this set is given by the direct sum
\[ \Fix\left(\Gamma_\z\right) \vee_{\mathrm{fix}} \Fix\left(\Gamma_{\z'}\right) = \Fix\left(\Gamma_\z\right) \oplus \Fix\left(\Gamma_{\z'}\right). \]
The meet operation is given by intersection
\[ \Fix\left(\Gamma_\z\right) \wedge_{\mathrm{fix}} \Fix\left(\Gamma_{\z'}\right) = \Fix\left(\Gamma_\z\right) \cap \Fix\left(\Gamma_{\z'}\right). \]

To be fully precise, one typically considers the lattice of \emph{conjugacy classes} of isotropy subgroups: Two subgroups $H,H'\le\Gamma$ are conjugate if there exists an element $\gamma\in\Gamma$ such that $\gamma^{-1}H\gamma=H'$. The partial order arises in terms of the subgroup relation of suitable representatives of the conjugacy classes. This lattice is called the \emph{isotropy lattice}. In that sense, it emerges from the above construction by summarizing all conjugate isotropy subgroups into their respective conjugacy class. In the same way, the fixed point subspaces can be summarized into classes of fixed point subspaces with conjugate isotropy subgroups. Both of these reduced lattices are in fact finite. For full details on the theoretical background, we refer to~\cite{Golubitsky.1988}~§10.
\clearpage
\subsection{Further Examples}
\subsubsection{Symmetries of Configurations}
\begin{example}
    \label{eq:triangle_app}
    Consider again the robot swarm in \Cref{fig:triangle} in which three robots are arranged in an equilateral triangle. 
    Let $\rho$  be the reflection through the indicated axis. In a suitable coordinate system this corresponds to a map on $\R^2$ which maps \emph{any point} $z=(x,y)$ to $(-x,y)$. This is not restricted to the positions of the three robots in the configuration. The induced map $M_\rho$ is a transformation of $\R^{2n}$. Applying this to the tuple of the configuration $\z$, one obtains
    \[ M_\rho \z = ((-x_1,y_1),(-x_2,y_2),(-x_3,y_3)). \]
    Due to the particular placement of $z_1,z_2,z_3$, one observes $x_1=0$ and $-x_2=x_3$ (or equivalently $\rho z_1=z_1$ and $\rho z_2=z_3$ and vice versa). In particular,
    \[ M_\rho \z = (z_1,z_3,z_2). \]
    Note that this tuple does not equal the original tuple $\z$. Instead, the final two entries are permuted. Consequently, if $\kappa\in S_3$ is the transposition of $2$ and $3$ one obtains
    \[ M_\kappa M_\rho \z = M_\kappa(z_1,z_3,z_2) = (z_1,z_2,z_3) = \z. \]
    Hence, $M_\kappa M_\rho \in \Gamma_\z$ is a symmetry of the configuration.

    By a similar consideration, if $\rho'$ is the rotation of $\R^2$ by $\frac{2\pi}{3}$ one observes that $\rho'z_1=z_2, \rho'z_2=z_3, \rho'z_3=z_1$. Let further, $\kappa'$ be the cyclic permutation shifting the indices by $1$. Then
    \[ M_{\kappa'} M_{\rho'} \z = M_{\kappa'}(z_2,z_3,z_1) = (z_1,z_2,z_3) = \z. \]
    Hence, also $M_{\kappa'} M_{\rho'} \in \Gamma_\z$ is a symmetry of the configuration. In particular, $\Gamma_\z$ consists of all possible products and powers of $M_\kappa M_\rho$ and $M_{\kappa'} M_{\rho'}$ where the matrices are explicitly given by
    \begin{align*}
        M_\kappa &= \begin{pmatrix}
            1 & 0 & 0 & 0 & 0 & 0 \\
            0 & 1 & 0 & 0 & 0 & 0 \\
            0 & 0 & 0 & 0 & 1 & 0 \\
            0 & 0 & 0 & 0 & 0 & 1 \\
            0 & 0 & 1 & 0 & 0 & 0 \\
            0 & 0 & 0 & 1 & 0 & 0 \\
        \end{pmatrix},
        &M_{\rho} &= \begin{pmatrix}
            -1 & 0 & 0 & 0 & 0 & 0 \\
            0 & 1 & 0 & 0 & 0 & 0 \\
            0 & 0 & -1 & 0 & 0 & 0 \\
            0 & 0 & 0 & 1 & 0 & 0 \\
            0 & 0 & 0 & 0 & -1 & 0 \\
            0 & 0 & 0 & 0 & 0 & 1 \\
        \end{pmatrix},\\
        M_{\kappa'} &= \begin{pmatrix}
            0 & 0 & 1 & 0 & 0 & 0 \\
            0 & 0 & 0 & 1 & 0 & 0 \\
            0 & 0 & 0 & 0 & 1 & 0 \\
            0 & 0 & 0 & 0 & 0 & 1 \\
            1 & 0 & 0 & 0 & 0 & 0 \\
            0 & 1 & 0 & 0 & 0 & 0 \\
        \end{pmatrix},
        &M_{\rho'} &= \begin{pmatrix}
            \frac{1}{2} & -\frac{\sqrt{3}}{2} & 0 & 0 & 0 & 0 \\
            \frac{\sqrt{3}}{2} & \frac{1}{2} & 0 & 0 & 0 & 0 \\
            0 & 0 & \frac{1}{2} & -\frac{\sqrt{3}}{2} & 0 & 0 \\
            0 & 0 & \frac{\sqrt{3}}{2} & \frac{1}{2} & 0 & 0 \\
            0 & 0 & 0 & 0 & \frac{1}{2} & -\frac{\sqrt{3}}{2} \\
            0 & 0 & 0 & 0 & \frac{\sqrt{3}}{2} & \frac{1}{2} \\
        \end{pmatrix}. \quad
    \end{align*}
\end{example}
\clearpage
\subsubsection{Explicit Lattice Derivations}
\begin{example}
    \label{ex:increase}
    In this example, we illustrate the theoretical contents of this section explicitly. Consider a swarm of $n=6$ robots. We assume that its current configuration $\z\in \R^{2n}$ is such that the robots are arranged in two concentric equilateral triangles (\Cref{fig:increase00}).
    \begin{figure}[!htb]
    \centering
    \resizebox{.3\linewidth}{!}{%
            \input{pics/full_example_00.tex}
        }
    \caption{Robot swarm consisting of $6$ robots discussed in~\Cref{ex:increase}. Gray lines indicate the initial arrangement into two concentric equilateral triangles. Red lines indicate the potential identification as a star shape (cf.~\Cref{fig:preservation}). Labels are assigned arbitrarily.}
    \label{fig:increase00}
    \end{figure}
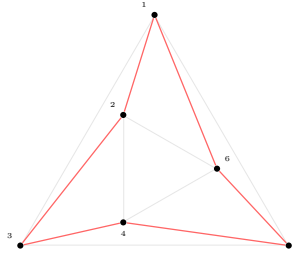    
    We want to use the algebraic considerations to elucidate all potential symmetries that any future configuration $\z^+$ might have or that any previous configuration $\z^-$ might have had when the swarm evolves due to a protocol according to the \fsync{} \oblot/ model. More precisely, we want to derive that part of the symmetry lattice~\eqref{eq:lattice} that is up- or downstream connected to $\Gamma_\z$. Similar to \Cref{ex:triangle} (\ref{triangle_item}), the isotropy subgroup of $\z\in \R{^2n}$ is
    \[ \Gamma_\z = \left\langle M_\kappa M_\rho \right\rangle, \]
    where $\rho$ is the rotation by $\frac{2\pi}{3}$ and the permutation is given by
    \[ \kappa (1,2,3,4,5,6) = (3,4,5,6,1,2). \]
    Note that this subgroup contains three rotations but no reflections.

    We begin by investigating the potential symmetries that the previous configuration $\z^-$ might have had. To that end, we consider all possible subgroups of $\Gamma_\z$. These arise by removing elements from $\Gamma_\z$ such that the remaining set is still closed under multiplication. Since $3$ is prime, there are no non-trivial possibilities. Hence, the only subgroup is $\set{\id_{2^n}}$ which is the isotropy group of any non-symmetric swarm.

    To deduce possibilities for gaining symmetries, we have to investigate supergroups $\Gamma_{\z^+}\ge\Gamma_\z$. Enlarging the isotropy group $\Gamma_\z$ means adding at least one additional element $\gamma\in\Gamma$ so that
    \[ \Gamma_{\z^+} \ge \left\langle M_\kappa M_\rho, \gamma \right\rangle. \]
    The ways in which a super \emph{isotropy} group can emerge is illustrated in \Cref{fig:increase} and can be listed as follows.

     \begin{figure}[!htb]
    \centering
    \begin{tabular}{ccc}
        \subfloat[Reflection symmetry \[\Gamma_{\z^+}=\left\langle M_\kappa M_\rho, M_{\kappa'}M_{\rho'}\right\rangle.\]]{
            \label{fig:increase04}
        \resizebox{.3\linewidth}{!}{%
            \input{pics/full_example_04.tex}
        }
        } &
        \subfloat[Reflection symmetry \[\Gamma_{\z^+}=\left\langle M_\kappa M_\rho, M_{\kappa''}M_{\rho''}\right\rangle.\]]{
            \label{fig:increase05}
        \resizebox{.3\linewidth}{!}{%
            \input{pics/full_example_05.tex}
        }
        } &
        \subfloat[Rotational symmetry \begin{multline*}
            \Gamma_{\z^+} = \left\langle M_{\kappa}M_{\rho}, M_{\overline\kappa}M_{\overline\rho}, \right. \\ \left. M_{\kappa'}M_{\rho'}, M_{\kappa''}M_{\rho''} \right\rangle,
        \end{multline*}]{
            \label{fig:increase06}
        \resizebox{.3\linewidth}{!}{%
            \input{pics/full_example_06.tex}
        }
        } \\
        \subfloat[Permutation and reflection symmetry \[\Gamma_{\z^+}=\left\langle M_\kappa M_\rho, M_{\tilde\kappa_1}, M_{\kappa''}M_{\rho''}\right\rangle.\]]{
            \label{fig:increase01}
        \resizebox{.3\linewidth}{!}{%
            \input{pics/full_example_01.tex}
        }
        } &
        \subfloat[Permutation and reflection symmetry \[\Gamma_{\z^+}=\left\langle M_\kappa M_\rho, M_{\tilde\kappa_2}, M_{\kappa''}M_{\rho''}\right\rangle.\]]{
            \label{fig:increase02}
        \resizebox{.3\linewidth}{!}{%
            \input{pics/full_example_02.tex}
        }
        } &
        \subfloat[Permutation and reflection symmetry \[\Gamma_{\z^+}=\left\langle M_\kappa M_\rho, M_{\tilde\kappa_3}, M_{\kappa^*_3}M_{\rho^*_3}\right\rangle.\]]{
            \label{fig:increase03}
        \resizebox{.3\linewidth}{!}{%
            \input{pics/full_example_03.tex}
        }
        }
    \end{tabular}
    \caption{Illustration of potential and future  symmetries of the robot swarm in~\Cref{fig:increase00}. For a detailed discussion including the lattice of isotropy subgroups, see~\Cref{ex:increase}.}
    \label{fig:increase}
\end{figure}
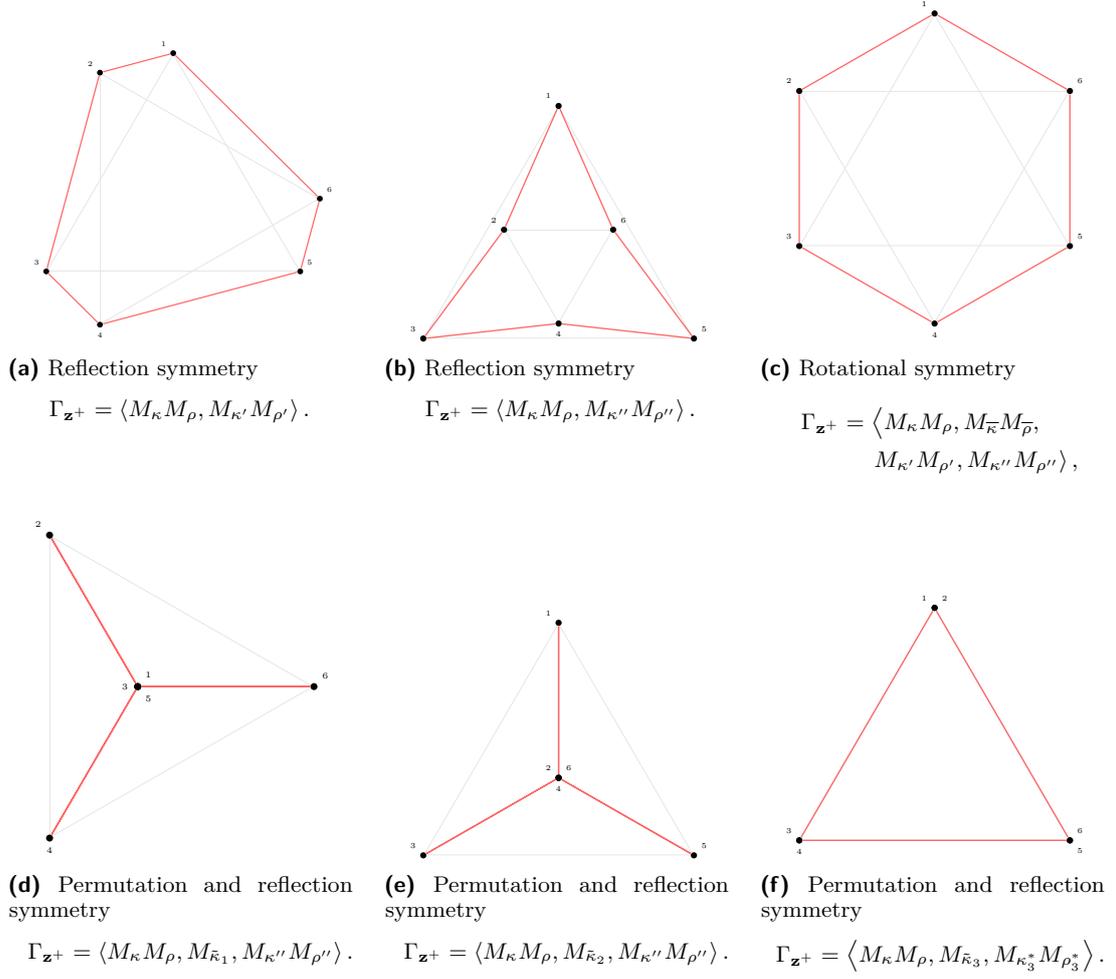

    \begin{itemize}
        \item An additional reflection symmetry $\tilde\rho$ without rearranging labels, i.e., $\gamma=M_{\tilde\rho}$, requires $\z^+$ to be collinear on the reflection axis. Due to the rotational symmetries $M_{\kappa}M_{\rho}$ this has to be true for all three rotated images of the reflection axis as well. This can only be true if all robots are in the center of the rotation. In particular, all robots have collided, a situation which is fixed under all potential symmetries which yields
        \[ \Gamma_{\z^+}=\Gamma. \]
        \item If $\gamma=M_{\tilde\rho}$ for some rotation $\tilde\rho$ (without rearranging labels) the configuration has to be fixed under a rotation without rearranging the labels. This can only be true if all robots are in the rotation center. As before, this yields
        \[ \Gamma_{\z^+}=\Gamma. \]
        \item If $\gamma$ is the combination of a reflection with a corresponding rearrangement of the labels in almost all cases the larger isotropy subgroup is the entire group $\Gamma$ by similar  mechanisms as before (robots being forced to be on multiple reflection axes or in rotation centers). Effectively, there are two exceptions:
        \begin{itemize}
            \item The inner triangle becomes a reflection of the outer which adds three rotationally related reflection symmetries (the specifics of which depend on which robots are reflections of each other). This situation  requires both triangles in $\z^+$ to have the same size. The configuration in~\Cref{fig:increase04} has the additional symmetry $\gamma=M_{\kappa'}M_{\rho'}$, where $\rho'$ is the reflection through the axis through the midpoints between $z_1$ and $z_2$ and that between robots $z_4$ and $z_5$ and
            \[ \kappa'(1,2,3,4,5,6)=(2,1,6,5,4,3). \]
            \item The inner triangle evolves such that each of its robots has the same distance to precisely two robots on the outer triangle. This adds three rotationally related reflection axes through one inner and one outer robot each. For example, the configuration in~\Cref{fig:increase05} has the additional symmetry $\gamma=M_{\kappa''}M_{\rho''}$, where $\rho''$ is the reflection through the axis through $z_1$ and $z_4$ and
            \[ \kappa''(1,2,3,4,5,6)=(1,6,5,4,3,2). \]
        \end{itemize}
        Any other reflection symmetry either automatically requires early collisions or multiple added symmetries as described in the next bullet point.
        \item Similarly, if $\gamma$ is the combination of a rotation with a corresponding rearrangement of the labels in almost all cases the larger isotropy subgroup is the entire group $\Gamma$. The only exception is an additional rotational symmetry by half the previous rotation angle. That is, $\gamma=M_{\overline\kappa}M_{\overline\rho}$, where $\overline{\rho}$ is the rotation by $\frac{1}{2}\cdot\frac{2\pi}{3}=\frac{\pi}{3}$ and
        \[ \overline{\kappa}(1,2,3,4,5,6)=(2,3,4,5,6,1). \]
        In this case, the configuration is fixed under rotations by half the previous rotation angle (after suitable rearranging the labels) which requires the two triangles to have merged into a regular $6$-gon (\Cref{fig:increase06}). In this case, the configuration is also invariant under both of the reflections discussed in the previous bullet point (probably with a different relabeling). In particular, this yields
        \[ \Gamma_{\z^+} = \left\langle M_{\kappa}M_{\rho}, M_{\overline\kappa}M_{\overline\rho}, M_{\kappa'}M_{\rho'}, M_{\kappa''}M_{\rho''} \right\rangle, \]
        which trivially contains the previous two isotropy subgroups as intermediate steps.
        \item If $\gamma=M_{\tilde\kappa}$ for some permutation $\tilde\kappa$ the evolved configuration is such that the robots with non-trivially permuted labels (via $\tilde\kappa$) must be in the same position. This permutation $\tilde\kappa$ further interacts with $\kappa$ to add additional permutations to $\Gamma_{\z^+}$. Four possibilities can be obtained
        \begin{itemize}
            \item The added permutation $\tilde\kappa_1$ leads to permutations of $1,3,5$ only. These require robots $1,3,5$ to have collided. In this case, the configuration automatically acquires reflection symmetries. These  are the three rotations of $M_{\kappa''}M_{\rho''}$, from before (\Cref{fig:increase01}). Hence,
            \[ \Gamma_\z^+=\left\langle M_\kappa M_\rho, M_{\tilde\kappa_1}, M_{\kappa''}M_{\rho''}\right\rangle. \]
            In particular, only adding the reflection symmetry is an intermediate step.
            \item The added permutation $\tilde\kappa_2$ leads to permutations of $2,4,6$ only. These require robots $2,4,6$ to have collided. In this case, the configuration automatically acquires reflection symmetries. These  are the three rotations of $M_{\kappa''}M_{\rho''}$, from before (\Cref{fig:increase02}). Hence,
            \[ \Gamma_\z^+=\left\langle M_\kappa M_\rho, M_{\tilde\kappa_1}, M_{\kappa''}M_{\rho''}\right\rangle. \]
            In particular, only adding the reflection symmetry is an intermediate step.
            \item The added permutation $\tilde\kappa_3$ leads to transpositions of $1,2$ and $3,4$ and $5,6$. These require the two triangles to have collided such that only one remains. In this case, the configuration automatically acquires reflection symmetries. These  are the three rotations of $M_{\kappa^*}M_{\rho^*}$, where $\rho^*$ is the reflection through the axis through $z_1$ and the midpoint between $z_3$ and $z_5$ and (for example depending on the precise collisions)
            \[ \kappa^*_1(1,2,3,4,5,6)=(1,2,5,6,3,4) \]
            (\Cref{fig:increase03}). Note that this is a different reflection as that induced by $\rho''$, as only one robot of each triangle is on the reflection axis. Hence,
            \[ \Gamma_{\z^+}=\left\langle M_\kappa M_\rho, M_{\tilde\kappa_3}, M_{\kappa^*_3}M_{\rho^*_3}\right\rangle. \]
            \item The added permutation leads to permutations  of arbitrary labels. These require all robots to have collided in which case
            \[ \Gamma_{\z^+}=\Gamma, \]
            as before. This isotropy subgroup contains all previous cases as intermediate steps.
        \end{itemize}
    \end{itemize}
    In any of these cases, a further gain of symmetries in a subsequent step can only happen by evolving into a fully symmetric configuration, i.e., collision of all robots. To summarize, the relevant part of the symmetry lattice can be illustrated as follows.
    \begin{equation*}
    \hspace{-3em}
    \begin{tikzcd}[every arrow/.append style={dash}]
        &&&\Gamma \ar[dll]\ar[dl]\ar[d]\ar[dr]&\\
        & \begin{tabular}{c}
              $\left\langle M_{\kappa}M_{\rho}, M_{\overline\kappa}M_{\overline\rho},\right.$ \\
             $\left.M_{\kappa'}M_{\rho'}, M_{\kappa''}M_{\rho''} \right\rangle$
        \end{tabular}
          \ar[d]\ar[dr]
            &\begin{tabular}{c}
                $\left\langle M_\kappa M_\rho, M_{\tilde\kappa_1},\right.$ \\
             $\left. M_{\kappa''}M_{\rho''}\right\rangle$
            \end{tabular}\ar[d]
            &\begin{tabular}{c}
                $\left\langle M_\kappa M_\rho, M_{\tilde\kappa_2},\right.$ \\
             $\left. M_{\kappa''}M_{\rho''}\right\rangle$
            \end{tabular} \ar[dl] 
            &\begin{tabular}{c}
                $\left\langle M_\kappa M_\rho, M_{\tilde\kappa_3},\right.$ \\
             $\left. M_{\kappa^*}M_{\rho^*}\right\rangle$
            \end{tabular} \ar[ddl]& \\
        & \left\langle M_\kappa M_\rho, M_{\kappa'}M_{\rho'}\right\rangle\ar[drr]
            &\left\langle M_\kappa M_\rho, M_{\kappa''}M_{\rho''}\right\rangle \ar[dr]
            && \\
        &&&\Gamma_\z=\left\langle M_\kappa M_\rho \right\rangle \ar[d]& \\
        &&&\set{\id_{2n}}&
    \end{tikzcd}
    \end{equation*}
\end{example}

\end{document}

%% file: pics/full_example_00.tex
\begin{tikzpicture}[
    		swarmx/.style = {
    			circle,
    			fill={black},
    			opacity=1,
    			inner sep=1pt
    		},
    		poly/.style = {
    	   		regular polygon,
    			regular polygon sides=3,
    			draw,
    			thin,
    			color=gray!20
    		}
    		]
    		\node[poly, inner sep=25] (p1) {};
    		\node[swarmx,label={[font=\tiny,scale=.7]135:$1$}] (1) at (p1.corner 1) {};
    		\node[swarmx,label={[font=\tiny,scale=.7]135:$3$}] (2) at (p1.corner 2) {};
    		\node[swarmx,label={[font=\tiny,scale=.7]45:$5$}] (3) at (p1.corner 3) {};
    	
    		\node[poly, inner sep=10, rotate=30] (p2) {};
    		\node[swarmx,label={[font=\tiny,scale=.7]135:$2$}] (8) at (p2.corner 1) {};
    		\node[swarmx,label={[font=\tiny,scale=.7]270:$4$}] (9) at (p2.corner 2) {};
    		\node[swarmx,label={[font=\tiny,scale=.7]45:$6$}] (10) at (p2.corner 3) {};
    		
            \draw[red!60, line width=.6pt] (1) to (8) to (2) to (9) to (3) to (10) to (1);
        
\end{tikzpicture}

%% file: pics/full_example_04.tex
\begin{tikzpicture}[
    		swarmx/.style = {
    			circle,
    			fill={black},
    			opacity=1,
    			inner sep=1pt
    		},
    		poly/.style = {
    	   		regular polygon,
    			regular polygon sides=3,
    			draw,
    			thin,
    			color=gray!20
    		}
    		]
    		\node[poly, inner sep=25] (p1) {};
    		\node[swarmx,label={[font=\tiny,scale=.7]135:$1$}] (1) at (p1.corner 1) {};
    		\node[swarmx,label={[font=\tiny,scale=.7]135:$3$}] (2) at (p1.corner 2) {};
    		\node[swarmx,label={[font=\tiny,scale=.7]45:$5$}] (3) at (p1.corner 3) {};
    	
    		\node[poly, inner sep=25, rotate=30] (p2) {};
    		\node[swarmx,label={[font=\tiny,scale=.7]135:$2$}] (8) at (p2.corner 1) {};
    		\node[swarmx,label={[font=\tiny,scale=.7]270:$4$}] (9) at (p2.corner 2) {};
    		\node[swarmx,label={[font=\tiny,scale=.7]45:$6$}] (10) at (p2.corner 3) {};
    		
            \draw[red!60, line width=.6pt] (1) to (8) to (2) to (9) to (3) to (10) to (1);
            
\end{tikzpicture}

%% file: pics/full_example_05.tex
\begin{tikzpicture}[
    		swarmx/.style = {
    			circle,
    			fill={black},
    			opacity=1,
    			inner sep=1pt
    		},
    		poly/.style = {
    	   		regular polygon,
    			regular polygon sides=3,
    			draw,
    			thin,
    			color=gray!20
    		}
    		]
    		\node[poly, inner sep=25] (p1) {};
    		\node[swarmx,label={[font=\tiny,scale=.7]135:$1$}] (1) at (p1.corner 1) {};
    		\node[swarmx,label={[font=\tiny,scale=.7]135:$3$}] (2) at (p1.corner 2) {};
    		\node[swarmx,label={[font=\tiny,scale=.7]45:$5$}] (3) at (p1.corner 3) {};
    	
    		\node[poly, inner sep=10, rotate=60] (p2) {};
    		\node[swarmx,label={[font=\tiny,scale=.7]135:$2$}] (8) at (p2.corner 1) {};
    		\node[swarmx,label={[font=\tiny,scale=.7]270:$4$}] (9) at (p2.corner 2) {};
    		\node[swarmx,label={[font=\tiny,scale=.7]45:$6$}] (10) at (p2.corner 3) {};
    		
            \draw[red!60, line width=.6pt] (1) to (8) to (2) to (9) to (3) to (10) to (1);
            
\end{tikzpicture}

%% file: pics/full_example_06.tex
\begin{tikzpicture}[
    		swarmx/.style = {
    			circle,
    			fill={black},
    			opacity=1,
    			inner sep=1pt
    		},
    		poly/.style = {
    	   		regular polygon,
    			regular polygon sides=3,
    			draw,
    			thin,
    			color=gray!20
    		}
    		]
    		\node[poly, inner sep=25] (p1) {};
    		\node[swarmx,label={[font=\tiny,scale=.7]135:$1$}] (1) at (p1.corner 1) {};
    		\node[swarmx,label={[font=\tiny,scale=.7]135:$3$}] (2) at (p1.corner 2) {};
    		\node[swarmx,label={[font=\tiny,scale=.7]45:$5$}] (3) at (p1.corner 3) {};
    	
    		\node[poly, inner sep=25, rotate=60] (p2) {};
    		\node[swarmx,label={[font=\tiny,scale=.7]135:$2$}] (8) at (p2.corner 1) {};
    		\node[swarmx,label={[font=\tiny,scale=.7]270:$4$}] (9) at (p2.corner 2) {};
    		\node[swarmx,label={[font=\tiny,scale=.7]45:$6$}] (10) at (p2.corner 3) {};
    		
            \draw[red!60, line width=.6pt] (1) to (8) to (2) to (9) to (3) to (10) to (1);
            
\end{tikzpicture}

%% file: pics/full_example_01.tex
\begin{tikzpicture}[
    		swarmx/.style = {
    			circle,
    			fill={black},
    			opacity=1,
    			inner sep=1pt
    		},
    		poly/.style = {
    	   		regular polygon,
    			regular polygon sides=3,
    			draw,
    			thin,
    			color=gray!20
    		}
    		]
            \node[swarmx,label={[font=\tiny,scale=.7]60:$1$}] (1) {};
    		\node[swarmx,label={[font=\tiny,scale=.7]180:$3$}] (2) {};
    		\node[swarmx,label={[font=\tiny,scale=.7]300:$5$}] (3) {};
    	
    		\node[poly, inner sep=25, rotate=30] (p2) {};
    		\node[swarmx,label={[font=\tiny,scale=.7]135:$2$}] (8) at (p2.corner 1) {};
    		\node[swarmx,label={[font=\tiny,scale=.7]270:$4$}] (9) at (p2.corner 2) {};
    		\node[swarmx,label={[font=\tiny,scale=.7]45:$6$}] (10) at (p2.corner 3) {};
    		
            \draw[red!60, line width=.6pt] (1) to (8) to (2) to (9) to (3) to (10) to (1);
            
\end{tikzpicture}

%% file: pics/full_example_02.tex
\begin{tikzpicture}[
    		swarmx/.style = {
    			circle,
    			fill={black},
    			opacity=1,
    			inner sep=1pt
    		},
    		poly/.style = {
    	   		regular polygon,
    			regular polygon sides=3,
    			draw,
    			thin,
    			color=gray!20
    		}
    		]
    		\node[poly, inner sep=25] (p1) {};
    		\node[swarmx,label={[font=\tiny,scale=.7]135:$1$}] (1) at (p1.corner 1) {};
    		\node[swarmx,label={[font=\tiny,scale=.7]135:$3$}] (2) at (p1.corner 2) {};
    		\node[swarmx,label={[font=\tiny,scale=.7]45:$5$}] (3) at (p1.corner 3) {};
    		\node[swarmx,label={[font=\tiny,scale=.7]135:$2$}] (8) {};
    		\node[swarmx,label={[font=\tiny,scale=.7]270:$4$}] (9) {};
    		\node[swarmx,label={[font=\tiny,scale=.7]45:$6$}] (10) {};
    		
            \draw[red!60, line width=.6pt] (1) to (8) to (2) to (9) to (3) to (10) to (1);
            
\end{tikzpicture}

%% file: pics/full_example_03.tex
\begin{tikzpicture}[
    		swarmx/.style = {
    			circle,
    			fill={black},
    			opacity=1,
    			inner sep=1pt
    		},
    		poly/.style = {
    	   		regular polygon,
    			regular polygon sides=3,
    			draw,
    			thin,
    			color=gray!20
    		}
    		]
    		\node[poly, inner sep=25] (p1) {};
    		\node[swarmx,label={[font=\tiny,scale=.7]135:$1$}] (1) at (p1.corner 1) {};
    		\node[swarmx,label={[font=\tiny,scale=.7]135:$3$}] (2) at (p1.corner 2) {};
    		\node[swarmx,label={[font=\tiny,scale=.7]315:$5$}] (3) at (p1.corner 3) {};
    	
    		\node[poly, inner sep=25, rotate=0] (p2) {};
    		\node[swarmx,label={[font=\tiny,scale=.7]45:$2$}] (8) at (p2.corner 1) {};
    		\node[swarmx,label={[font=\tiny,scale=.7]270:$4$}] (9) at (p2.corner 2) {};
    		\node[swarmx,label={[font=\tiny,scale=.7]45:$6$}] (10) at (p2.corner 3) {};
    		
            \draw[red!60, line width=.6pt] (1) to (2) to (3) to (1);
        
\end{tikzpicture}